\documentclass[twoside,11pt]{article}

\usepackage{jmlr}
\usepackage{amsmath}
\usepackage{subfigure}
\usepackage{algorithm}
\usepackage{algorithmic}
\usepackage{multirow}

\newtheorem{assumption}{Assumption}
\newtheorem{property}{Property}

\jmlrheading{1}{2000}{1-48}{4/00}{10/00}{meila00a}{Shen-Yi Zhao, Hao Gao and Wu-Jun Li}

\def\a{{\bf a}}

\def\e{{\bf e}}

\def\g{{\bf g}}

\def\u{{\bf u}}

\def\w{{\bf w}}
\def\x{{\bf x}}
\def\y{{\bf y}}

\def\A{{\bf A}}

\def\I{{\bf I}}

\def\P{{\bf P}}

\def\X{{\bf X}}

\def\0{{\bf 0}}
\def\1{{\bf 1}}
\def\2{{\bf 2}}
\def\3{{\bf 3}}
\def\4{{\bf 4}}
\def\5{{\bf 5}}
\def\6{{\bf 6}}
\def\7{{\bf 7}}
\def\8{{\bf 8}}
\def\9{{\bf 9}}

\def\EB{{\mathbb E}}

\def\RB{{\mathbb R}}

\begin{document}

\title{ADASS: Adaptive Sample Selection  for Training Acceleration}

\author{\name Shen-Yi Zhao \email zhaosy@lamda.nju.edu.cn \\
        \name Hao Gao      \email gaoh@lamda.nju.edu.cn \\
        \name Wu-Jun Li    \email liwujun@nju.edu.cn \\
        \addr Department of Computer Science and Technology\\
       Nanjing University, China}

\maketitle

\begin{abstract}
Stochastic gradient decent~(SGD) and its variants, including some accelerated variants, have become popular for training in machine learning. However, in all existing SGD and its variants, the sample size in each iteration~(epoch) of training is the same as the size of the full training set. In this paper, we propose a new method, called \underline{ada}ptive \underline{s}ample \underline{s}election~(ADASS), for training acceleration. During different epoches of training, ADASS only need to visit different training subsets which are adaptively selected from the full training set according to the Lipschitz constants of the loss functions on samples. It means that in ADASS the sample size in each epoch of training can be smaller than the size of the full training set, by discarding some samples. ADASS can be seamlessly integrated with existing optimization methods, such as SGD and momentum SGD, for training acceleration. Theoretical results show that the learning accuracy of ADASS is comparable to that of counterparts with full training set. Furthermore, empirical results on both shallow models and deep models also show that ADASS can accelerate the training process of existing methods without sacrificing accuracy.
\end{abstract}


\section{Introduction}
Many machine learning models can be formulated as the following empirical risk minimization problem:
\begin{align}\label{goal:erm}
	\mathop{\min}_{\w} F(\w) := \frac{1}{n}\sum_{i=1}^{n} f_i(\w),
\end{align}
where $\w$ is the parameter to learn, $f_i(\w)$ corresponds to the loss on the $i$th training sample, and $n$ is the total number of training samples.

With the rapid growth of data in real applications, stochastic optimization methods have become more popular than batch ones to solve the problem in~(\ref{goal:erm}). The most popular stochastic optimization method is stochastic gradient decent~(SGD)~\citep{DBLP:conf/icml/Zhang04,DBLP:conf/nips/Xiao09,bottou-2010,DBLP:conf/colt/DuchiHS10}. One practical way to adopt SGD for learning is the so-called Epoch-SGD~~\citep{DBLP:journals/jmlr/HazanK14} in Algorithm~\ref{alg:epochsgd}, which has been widely used by mainstream machine learning platforms like Pytorch and TensorFlow. In each outer iteration~(also called epoch) of Algorithm~\ref{alg:epochsgd}, Epoch-SGD first samples a sequence $\{i_1, i_2, \ldots,i_n\}$ from $\{1,2,\ldots,n\}$ according to a distribution $p_t$ defined on the full training set. A typical distribution is the uniform distribution. We can also set the sequence to be a permutation of $\{1,2,\ldots,n\}$~\citep{DBLP:journals/siamjo/Tseng98}. In the inner iteration, the stochastic gradients computed based on the sampled sequence will be used to update the parameter. The mini-batch size is one in the inner iteration of Algorithm~\ref{alg:epochsgd}. In real applications, larger mini-batch size can also be used. After the inner iteration is completed, Epoch-SGD adjusts the step size to guarantee that $\{\eta_t\}$ is a non-increasing sequence. In general, we take $\eta_{t+1} = \alpha \eta_t, \alpha \in (0,1)$. Although many theoretical results suggest $\w_{t+1}$ to be the average of $\{\u_{m}\}$, we usually take the last one $\u_{n}$ to be the initialization of the next outer iteration.

\begin{algorithm}[htb]
\caption{Epoch-SGD}
\label{alg:epochsgd}
\begin{algorithmic}[1]
\STATE Initialization: $\w_{1},\eta_1>0$;
\FOR{$t=1,2, \ldots,T$}
\STATE Let $\{i_1,i_2,\ldots,i_n\}$ be a sequence sampled from $\{1,2,\ldots,n\}$ according to a distribution $p_t$ defined on the full training set;
\STATE $\u_{0} = \w_{t}$;
\FOR{$m=1,2,\ldots,n$}
\STATE $\u_{m} = \u_{m-1} - \eta_t \nabla f_{i_m}(\u_{m-1})$;
\ENDFOR
\STATE $\w_{t+1} = \u_{n}$;
\STATE Adjust step size to get $\eta_{t+1}$;
\ENDFOR
\end{algorithmic}
\end{algorithm}

To further accelerate Epoch-SGD in Algorithm~\ref{alg:epochsgd}, three main categories of methods have recently been proposed. The first category is to adopt momentum, Adam or Nesterov's acceleration~\citep{Nesterov07gradientmethods,DBLP:conf/nips/LeenO93,DBLP:journals/siamjo/Tseng98,DBLP:journals/mp/Lan12,DBLP:journals/corr/KingmaB14,DBLP:journals/mp/GhadimiL16,DBLP:conf/icml/Allen-Zhu18} to modify the update rule of SGD in Line 6 of Algorithm~\ref{alg:epochsgd}. This category of methods has faster convergence rate than SGD when $t$ is small, and empirical results show that these methods are more stable than SGD. However, due to the variance of $\nabla f_{i_m}(\cdot)$, the convergence rate of these methods is the same as that in SGD when $t$ is large.

The second category is to design new stochastic gradients to replace $\nabla f_{i_m}(\cdot)$ in the inner iteration of Algorithm~\ref{alg:epochsgd} such that the variance in the stochastic gradients can be reduced~\citep{DBLP:conf/nips/Johnson013,DBLP:journals/jmlr/Shalev-Shwartz013,DBLP:conf/nips/Nitanda14,DBLP:conf/icml/Shalev-Shwartz014,DBLP:conf/nips/DefazioBL14,DBLP:journals/mp/SchmidtRB17}. Representative methods include SAG~\citep{DBLP:journals/mp/SchmidtRB17} and SVRG~\citep{DBLP:conf/nips/Johnson013}. These methods can achieve faster convergence rate than vanilla SGD in most cases. However, the faster convergence of these methods are typically based on a smooth assumption for the objective function, which might not be satisfied in real problems. Another disadvantage of these methods is that they usually need extra memory cost and computation cost to get the stochastic gradients.

The third category is the importance sampling based methods, which try to design the distribution $p_t$~\citep{DBLP:conf/icml/ZhaoZ15,DBLP:conf/icml/CsibaQR15,DBLP:conf/icml/NamkoongSYD17,DBLP:conf/icml/KatharopoulosF18,DBLP:conf/colt/Borsos0L18}. With properly designed distribution $p_t$, these methods can also reduce the variance of $\nabla f_{i_m}(\cdot)$ and hence achieve faster convergence rate than SGD.  \citep{DBLP:conf/icml/ZhaoZ15} designs a distribution according to the global Lipschitz or smoothness. The distribution is firstly calculated based on the training set and then is fixed during the whole training process. \citep{DBLP:conf/icml/CsibaQR15,DBLP:conf/icml/NamkoongSYD17} proposes an adaptive distribution which will change in each epoch. \citep{DBLP:conf/colt/Borsos0L18} adopts online optimization to get the adaptive distribution. There also exist some other heuristic importance sampling methods~\citep{DBLP:conf/cvpr/ShrivastavaGG16,DBLP:conf/iccv/LinGGHD17}, which mainly focus on training samples with large loss~(hard examples) and set the weight of samples with small loss to be small or $0$.

One shortcoming of SGD and its variants, including the accelerated variants introduced above, is that the sample size in each iteration~(epoch) of training is the same as the size of the full training set. This can also be observed in Algorithm~\ref{alg:epochsgd}, where a sequence of $n$ indices must be sampled from $\{1,2,\ldots,n\}$. Even for the importance sampling based methods, each sample in the full training set has possibility to be sampled in each outer iteration~(epoch) and hence no samples can be discarded during training.

In this paper, we propose a new method, called \underline{ada}ptive \underline{s}ample \underline{s}election~(ADASS), to solve the above shortcoming of existing SGD and its variants. The contributions of ADASS are outlined as follows:
\begin{itemize}
\item During different epoches of training, ADASS only need to visit different training subsets which are adaptively selected from the full training set according to the Lipschitz constants of the loss functions on samples. It means that in ADASS the sample size in each epoch of training can be smaller than the size of the full training set, by discarding some samples.
\item ADASS can be seamlessly integrated with existing optimization methods, such as SGD and momentum SGD, for training acceleration.
\item Theoretical results show that the learning accuracy of ADASS is comparable to that of counterparts with full training set.
\item Empirical results on both shallow models and deep models also show that ADASS can accelerate the training process of existing methods without sacrificing accuracy.
\end{itemize}

\section{Preliminary}
First, we give the following notations and definitions:
\begin{itemize}
  \item we use boldface lowercase letters like $\a$ to denote vectors, and use boldface uppercase letters
        like $\A$ to denote matrices;
  \item $\|\cdot\|$ denotes $L_2$ norm;
  \item $\e_i = (0,\ldots,1,\ldots,0)^T\in \RB^n$ with the $i$th element being $1$ and others being $0$;
  \item $[n] = \{1,2,\ldots,n\}$;
  \item $B(\w,r) = \{\w'| \|\w' - \w\| \leq r\}$;
\end{itemize}

\begin{definition}
	Let $\lambda > 0$. Function $\phi(\w)$ is called $\lambda$-strongly convex if $\forall \w, \w'$,
	\begin{align*}
		\phi(\w') \geq \phi(\w) + \nabla \phi(\w)^T(\w' - \w) + \frac{\lambda}{2}\|\w' - \w\|^2
	\end{align*}
\end{definition}

\begin{definition}
	Let $c > 0$. Function $\phi(\w)$ is called $c$-weakly convex if $\forall \w, \w'$,
	\begin{align*}
		\phi(\w') \geq \phi(\w) + \nabla \phi(\w)^T(\w' - \w) - \frac{c}{2}\|\w' - \w\|^2
	\end{align*}
\end{definition}

Second, we give some brief knowledge about SGD. It has been well known that
\begin{lemma}\label{lemma:SGD}
\citep{DBLP:conf/icml/Zinkevich03} Let $\phi(\w)$ be a convex function and $\Omega \in \RB^d$ be a convex domain, the sequence $\{\w_t\}$ is produced by
\begin{align*}
  \w_{t+1} = \mathop{\arg\min}_{\w\in \Omega} \g_t^T\w + \frac{1}{2\eta}\|\w - \w_t\|^2
\end{align*}
where $\EB[\g_t|\w_t] = \nabla \phi(\w_t), \EB\|\g_t\|^2 \leq G^2$ and $\eta$ is a constant. Let $\bar{\w}_T = \frac{1}{T}\sum_{t=0}^{T-1}\w_t$, then we have $\forall \w \in \Omega$,
\begin{align}\label{eq:converge_SGD}
  \EB [\phi(\bar{\w}_T) - \phi(\w)] \leq \frac{\|\w_0 - \w\|^2}{2T\eta} + \frac{G^2\eta}{2}.
\end{align}
\end{lemma}

In fact, most gradient based stochastic optimization methods, including SGD, momentum SGD, Adagrad, satisfy the following equation:
\begin{align}\label{eq:converge_SGD}
  \EB [\phi(\w_+) - \phi(\w)] \leq \theta_1\|\w_0 - \w\|^2 + \theta_2.
\end{align}
where $\w_0$ is the initialization, $\w_+$ is the output, $\theta_1, \theta_2$ are determined by the constant stepsize $\eta$, iterations $T$ and some other constants with respect to object $\phi(\cdot)$. We give out the $\theta_1, \theta_2$ of exact methods in the appendix.

\section{A Simple Case: Least Square}\label{sec:LS}

We first adopt least square to give some hints for designing effective sample selection strategies, because least square is a simple model with closed-form solution.

Given a training set $\{(\x_i,y_i)\}_{i=1}^n$, where $\x_i \in \RB^d$ and $y_i\in \RB$. Least square tries to optimize the following objective function:
\begin{align}\label{least square}
  \mathop{\min}_\w f(\w) := \frac{1}{2}\sum_{i=1}^{n}(y_i - \x_i^T\w)^2.
\end{align}
For convenience, let $\X = (\x_1,\x_2,\ldots,\x_n)\in \RB^{d\times n}, \y = (y_1,y_2,\ldots,y_n)^T\in \RB^{n\times 1}$. Furthermore, we assume $\mbox{rank}(\X\X^T) = d$, which is generally satisfied when $n\gg d$. Then, the optimal parameter $\w^*$ of~(\ref{least square}) can be directly computed as follows:
\begin{align}\label{eq:LinearR_optimum}
  \w^* = (\X\X^T)^{-1}\X\y .
\end{align}
Let $\{i_1,i_2,\ldots,i_n\}$ be a permutation of $\{1,2,\ldots,n\}$, and $\P_m = (\e_{i_1},\e_{i_2},\ldots,\e_{i_m})\in \RB^{n\times m}$. Then, $\X_m = \X\P_m$ and $\y_m = \P_m^T\y$ denote the features and supervised information of the $m$ selected samples indexed by $\{i_1,i_2,\ldots,i_m\}$. For simplicity, we assume $\mbox{rank}(\X\P_m\P_m^T\X^T) = d$. Then it is easy to get that
\begin{align}\label{eq:LinearR_suboptimum}
\w_m = & \mathop{\arg\min}_{\w}\sum_{k=1}^{m}(y_{i_k} - \x_{i_k}^T\w)^2 = (\X_m\X_m^T)^{-1}\X_m\y_m = (\X\A\P_m\P_m^T\X^T)^{-1}\X\P_m\P_m^T\y .
\end{align}
We are interested in the difference between $\w^*$ and $\w_m$. If the difference is very small, it means that we can use less training samples to estimate $\w$. We have the following lemma about the relationship between $\w^*$ and $\w_m$.
\begin{lemma}\label{lemma:LinearR_relation}
$\w^*$ and $\w_1$ satisfy the following equation:
\begin{align}
  (\X\P_m\P_m^T\X^T)(\w^* - \w_m) = \X(\I - \P_m\P_m^T)(\y - \X^T\w^*). \nonumber
\end{align}
\end{lemma}

Let $f_i(\w) = \frac{1}{2}(y_i - \x_i^T\w)^2$. Based on Lemma~\ref{lemma:LinearR_relation}, we can get the following theorem.
\begin{theorem}\label{coro:LinearR_bound}
Assume $\forall i, \|\nabla f_i(\w^*)\|\leq L_i, \|\x_i\|\leq a$. Let  $\{i_1,i_2,\ldots,i_n\}$ be a permutation of $\{1,2,\ldots,n\}$, and $\P_m = (\e_{i_1},\e_{i_2},\ldots,\e_{i_{m}})\in \RB^{n\times m}$. If $\exists m_0$ such that $rank(\X\P_{m_0}\P_{m_0}^T\X^T) = d$ with the smallest eigenvalue $b>0$, then $\forall m \geq m_0$, we have
\begin{align}
      &\|\w^* - \w_m\|^2 \leq (na^2/b^2)\sum_{k=m+1}^{n} f_{i_k}(\w^*) \nonumber \\
      &\|\w^* - \w_m\|^2 \leq (1/b^2)(\sum_{k=m+1}^{n} L_i)^2\nonumber
\end{align}
\end{theorem}

Here, $L_i$ actually corresponds to the Lipschitz constant of $f_i(\w)$ around $\w^*$. We call the bound of the first inequality in Theorem~\ref{coro:LinearR_bound} \emph{loss bound} because it is related to the loss on the samples. And we call the bound of the second inequality in Theorem~\ref{coro:LinearR_bound} \emph{Lipschitz bound} because it is related to the Lipschitz constants of the loss functions on the samples. We can find that in both loss bound and Lipschitz bound, the terms on the righthand side of the two inequalities correspond to those discarded~(un-selected) training samples indexed by $i_{m+1},\ldots,i_n$. Theorem~\ref{coro:LinearR_bound} gives us a hint that in least square, to make the gap between $\w^*$ and $\w_1$ as small as possible, we should discard $n-m$ training samples with the smallest losses or smallest Lipschitz constants. That means we should select $m$ training samples with the largest losses or largest Lipschitz constants.

\begin{figure}[t]
\centering
\subfigure[uniform distribution]{\includegraphics[width =4.5cm]{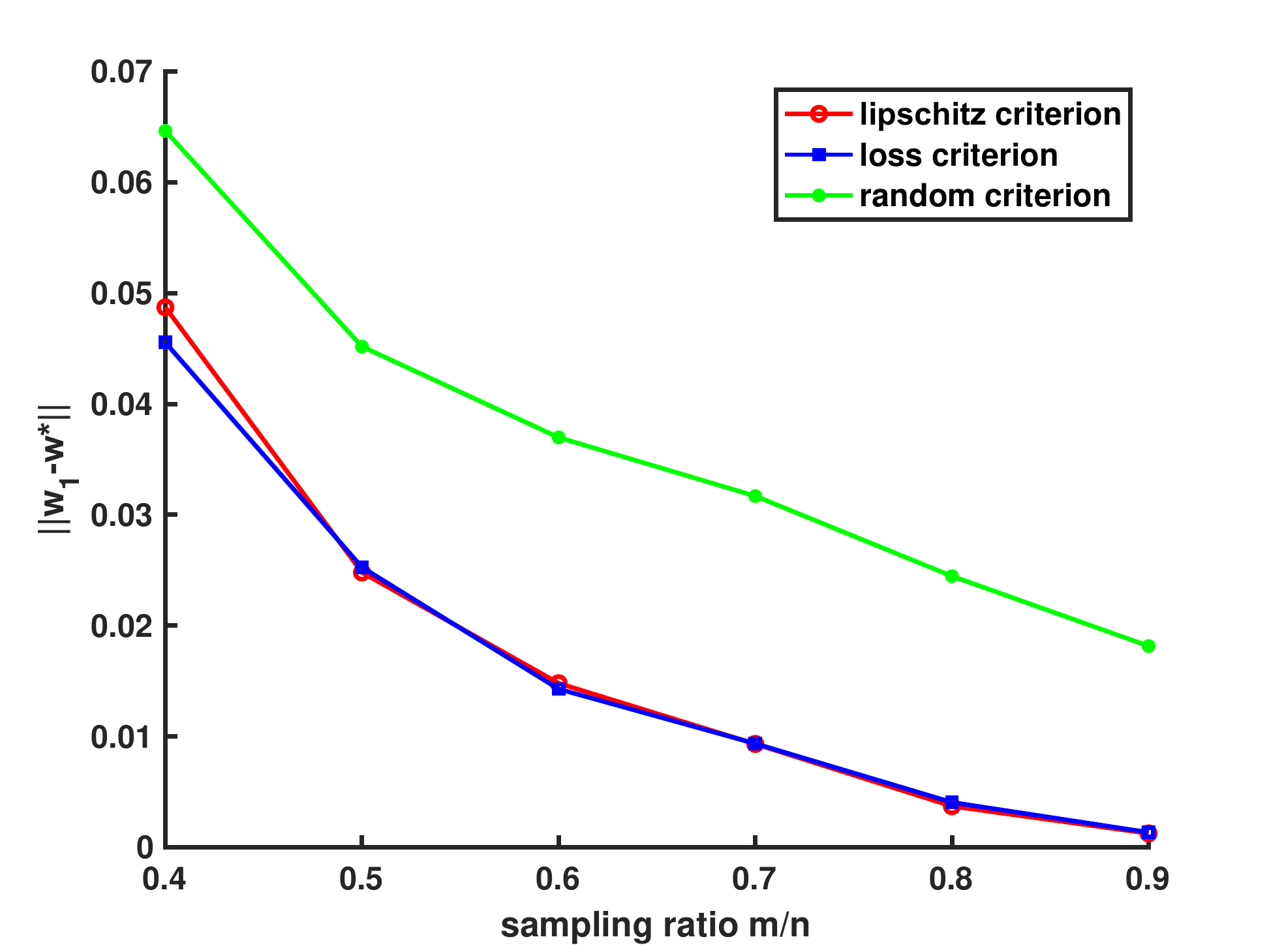}}
\subfigure[gaussian distribution]{\includegraphics[width =4.5cm]{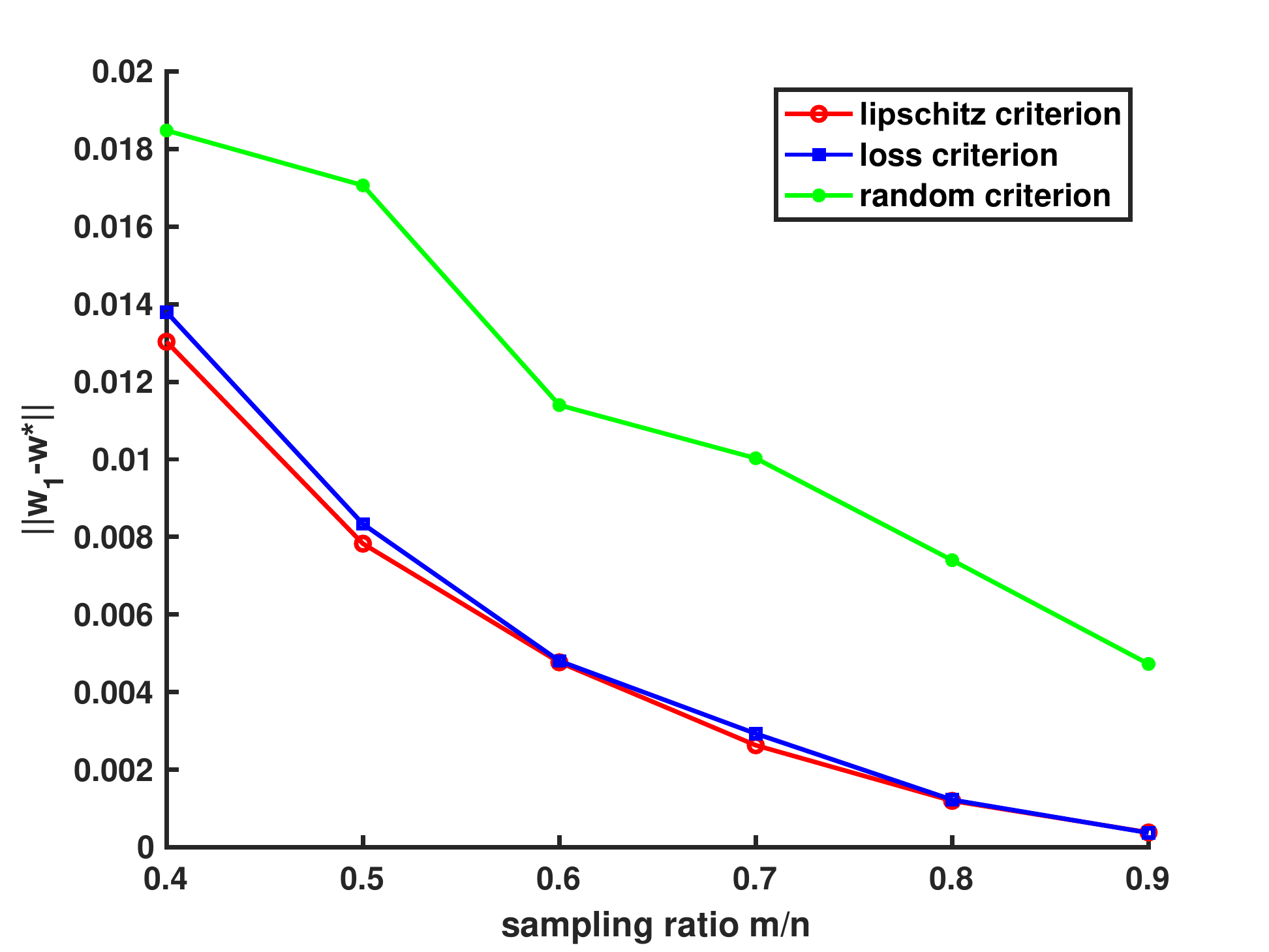}}
\subfigure[binomial distribution]{\includegraphics[width =4.5cm]{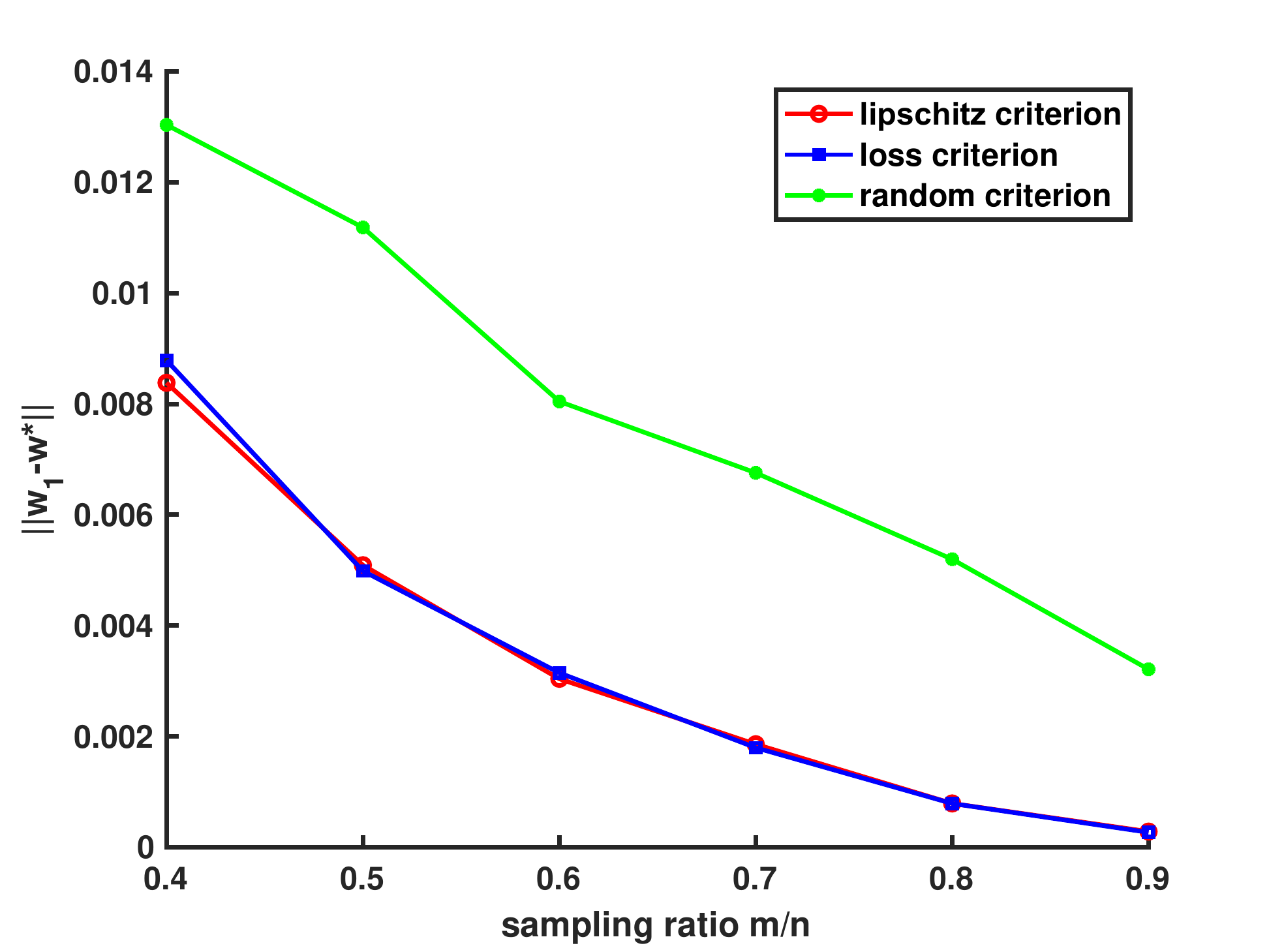}}
\caption{Empirical illustration to compare different sample section criterions for least square.}\label{exp:ls}
\end{figure}

We design an experiment to further illustrate the results in Theorem~\ref{coro:LinearR_bound}. The feature $\x_i$ is constructed from three different distributions: uniform distribution $\mathcal{U}(0,1)$, gaussian distribution $\mathcal{N}(0,1)$ and binomial distribution $\mathcal{B}(10,0.3)$. The corresponding $y_i$ is got by a linear transformation on $\x_i$ with a small gaussian noise. We compare three sample selection criterions: \emph{Lipschitz criterion} according to Lipschitz bound, \emph{loss criterion} according to loss bound, and \emph{random criterion} with which $m$ samples are randomly selected. The result is shown in Figure~\ref{exp:ls}, in which the y-axis denotes $\|\w^* - \w_m\|$ and the x-axis denotes the sampling ratio $m/n$. We can find that both Lipschitz criterion and loss criterion achieve better performance than random criterion, for estimating $\w^*$ with a subset of samples.

\section{Deep Analysis of Sample Selection Criterions}
\label{sec:ss}
Based on the results of Theorem~\ref{coro:LinearR_bound} about least square, it seems that both loss and Lipschitz constant can be adopted as criterions for sample selection. In this section, we give deep analysis about these two criterions and find that for general cases Lipschitz constant can be used for sample selection but loss cannot.

\subsection{Loss based Sample Selection}
\label{subsec:maxloss}

Based on the loss criterion, in each iteration, the algorithm will select samples with the largest loss at current $\w$ and learn with these selected samples to update $\w$. Intuitively, if the loss $f_i(\w)$ is large, it means the model has not fitted the $i$th sample good enough and this sample need to be trained again.

Unfortunately, the loss based sample selection cannot theoretically guarantee the convergence of the learning procedure. We can give a negative example as follows: let $n=2, f_1(w) = 1.5w, f_2(w) = -w+2, w\in [0,1], m = 1$. If we start from $w_0 = 0$, then we will get $w_1 = 1, w_2 = 0, w_3 = 1, \ldots$. It means $\{w_t\}$ is a divergent sequence. In fact, even $\w_{t+1}$ minimizes $\sum_{i\in S_t} f_i(\w)$ where $S_t$ is the $m$ selected samples with the largest loss at the $t$th iteration, it can also make the other un-selected sample loss functions increase.

The loss criterion has another disadvantage. Let $\mathbf{\epsilon} = (\epsilon_1,\epsilon_2,\ldots,\epsilon_n)$, and define
\begin{align}
	g_\epsilon(\w) = \sum_{i=1}^n (f_i(\w) + \epsilon_i). \nonumber
\end{align}
The $\mathbf{\epsilon}$ can be treated as some unknown noise. It is easy to find that $\forall \mathbf{\epsilon} \in \RB^n$,  minimizing $g_\epsilon(\w)$ is equivalent to minimizing $f(\w)$. However, $\epsilon$ can disrupt the samples in $S_t$ seriously. In Figure~\ref{exp:ls}, it is possible to design suitable $\epsilon$ to make the blue line be the same as the green line. Hence, the loss criterion is also not robust for sample selection.

\subsection{Lipschtiz Constant based Sample Selection}
\label{sec:ass}
In this subsection, we theoretically prove that Lipschtiz constant is a good criterion for sample selection. For convenience, we give the following notations: for any function $q(\w)$ and domain $\Omega$, we denote $\w_{q,\Omega}^* = \mathop{\arg\min}_{\w'\in \Omega} q(\w')$ and $\Delta(q,\w,\Omega) = q(\w) - q(\w_{q,\Omega}^*)$.

First, we give the following definition:
\begin{definition}\label{def:eos_insignificant}
Let $\phi_1(\cdot), \phi_2(\cdot)$ be two functions, $\Omega \subseteq \RB^d$ and $\w \in \Omega$. We say $\phi_2(\cdot)$ is $\zeta$-insignificant at $\w$ w.r.t $(\phi_1(\cdot),\Omega)$ if
\begin{align}
|\phi_2(\w_{\phi_1,\Omega}^*) - \phi_2(\w)| \leq \zeta(\phi_1(\w) -\phi_1(\w_{\phi_1,\Omega}^*)),  \label{eq:eos_insignificant}
\end{align}
\end{definition}
To further explain the definition, we consider the compositive function $h(\w) = \phi(\w) + \psi(\w)$. Then we have the following two properties:
\begin{property}\label{pro:1}
	If $\psi(\cdot)$ is $\zeta$-insignificant at $\w$ w.r.t $(\phi(\cdot),\Omega)$, $\zeta < 1$, then we have
	\begin{align*}
		h(\w_{\phi,\Omega}^*) \leq h(\w) - (1-\zeta)(\phi(\w) - \phi(\w_{\phi,\Omega}^*))
	\end{align*}
\end{property}
\begin{property}\label{pro:2}
	If $\psi(\cdot)$ is $\zeta$-insignificant at $\w$ w.r.t $(\phi(\cdot),\Omega)$, $\zeta < 1$, then we have
	\begin{align*}
		(1-\zeta)\Delta(\phi,\w,\Omega)\leq \Delta(h,\w,\Omega) \leq (1+\zeta)\Delta(\phi,\w,\Omega)
	\end{align*}
\end{property}

The first property implies that if $\psi(\cdot)$ is $\zeta$-insignificant, we can optimize $\phi(\cdot)$ on the domain $\Omega$ and the return $\w_{\phi,\Omega}^*$ will make the value of $h(\cdot)$ decrease, i.e. $h(\w_{\phi,\Omega}^*) \leq h(\w)$. The second property implies that if $\psi(\cdot)$ is $\zeta$-insignificant,  then optimizing $\phi(\cdot)$ is equivalent to optimizing $h(\cdot)$, i.e. $\Delta(h,\w,\Omega) = \Theta(\Delta(\phi,\w,\Omega))$. One trivial decomposition of $h(\w)$ is $\phi(\w) = \rho h(\w), \psi(\w)= (1-\rho)h(\w)$, where $\rho\in (0,1)$. Then $\psi(\cdot)$ is $\frac{1-\rho}{\rho}$-insignificant. In the following content, we are going to design a non-trivial decomposition of $h(\w)$ in which $\phi(\cdot)$ is easier to be optimized than $h(\cdot)$.

We assume $h(\w)$ has the structure of summation of $n$ functions, which means $h(\w) = \sum_{i=1}^n h_i(\w)$. We denote $h_S(\w) = \sum_{i\in S} h_i(\w), S\subseteq [n]$ and make the following assumption:
\begin{assumption}\label{ass:local lipschitz continuous}
(Local Lipschitz continuous)
$\forall \w\in \RB^d, r>0$, there exists a constant $L_i(\w,r)>0$ such that $\forall \w'\in B(\w,r)$,
\begin{align}
|h_i(\w') - h_i(\w)| \leq L_i(\w,r)\|\w' - \w\|.
\end{align}
\end{assumption}
For most loss functions used in machine learning, their gradients are bounded by a bounded closed domain which guarantees the Lipschitz continuous property. Hence, Assumption~\ref{ass:local lipschitz continuous} is satisfied by most machine learning models. $L_i(\w,r)$ is the local Lipschitz constant which is determined by the specific function parameter and the neighborhood size of $\w$. Hence, we set different Lipschitz constants for different $h_i(\cdot)$.
\begin{definition}\label{def:local strongly convex}
(Local one-point strongly convex)
Let $\w \in \Omega$. We say $h(\cdot)$ is \emph{$\mu$-local one-point strongly convex} at $\w$ if
\begin{align}
h_S(\w) - \mathop{\min}_{\w'\in \Omega}h_S(\w') \geq \frac{\mu|S|}{2}\|\w - \w_{h_S,\Omega}^*\|^2, \forall S\subseteq [n] \label{eq:weak}
\end{align}
\end{definition}
The 'local' is mainly for that (\ref{eq:weak}) holds for any subset $S$. Definition~\ref{def:local strongly convex} can also be satisfied by many machine learning models, which is explained as follows.
\begin{lemma}\label{lemma:strongconvex}
Assume each $h_i(\cdot)$ is $\mu$-strongly convex. Then $\forall \w$, $h(\cdot)$ is \emph{$\mu$-local one-point strongly convex} at $\w$.
\end{lemma}

Lemma~$\ref{lemma:strongconvex}$ implies that strongly convex objective functions has the local one-point strongly-convex property. For weakly convex objective functions, it is easy to transform them to strongly convex objective functions by adding a quadratic function.

\begin{figure}[!t]
\centering
  \includegraphics[width =5.5cm]{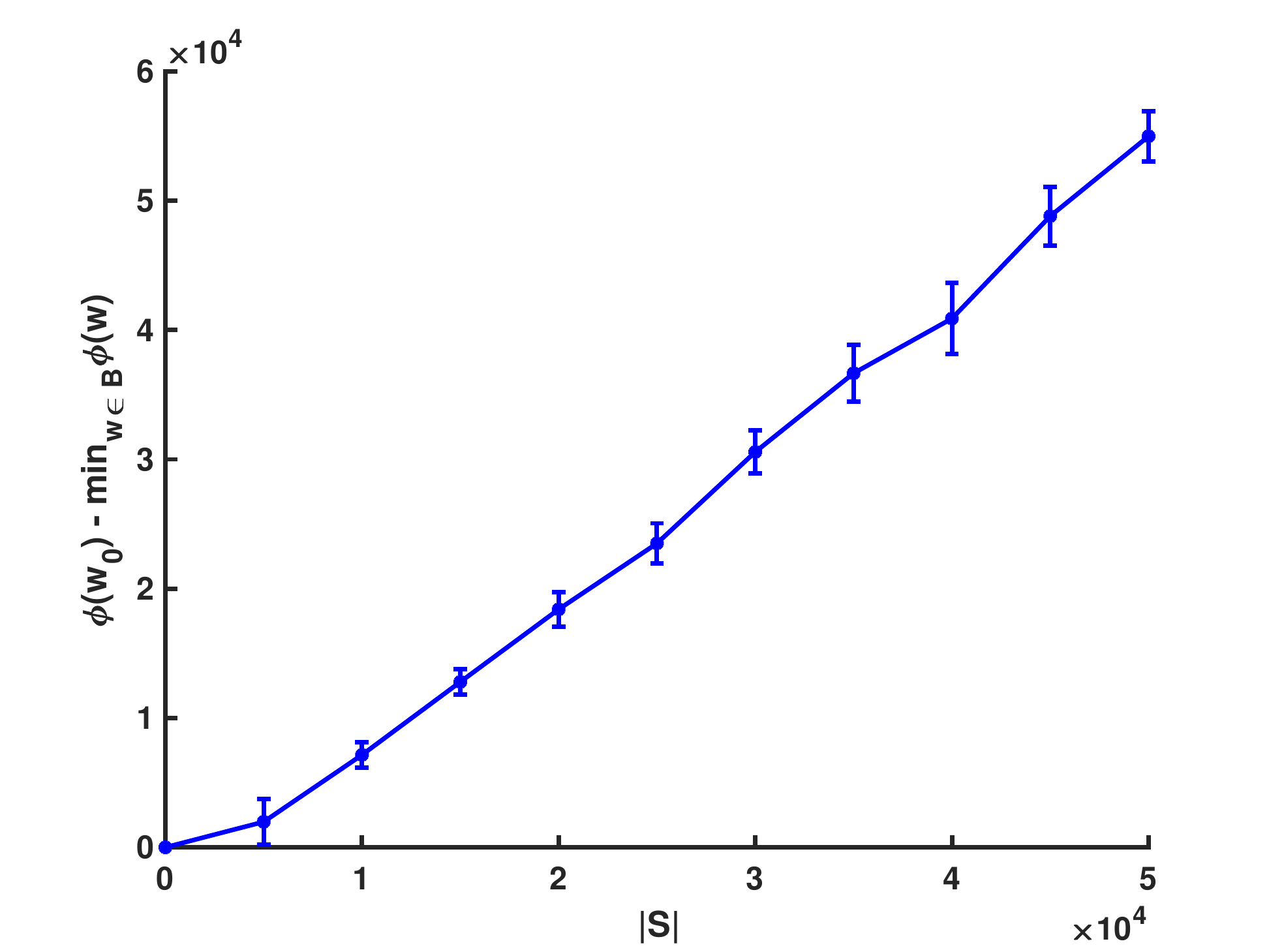}
  \caption{Empirical result for Definition~\ref{def:local strongly convex} on non-convex cases. The x-axis denotes $|S|$, which is the size of subset $S$. The y-axis represents the value $h_S(\w_0) - \mathop{\min}_{\w\in B(\w_0,r)}h_S(\w)$.}\label{exp:ass1}
\end{figure}

For non-convex objective functions, we also observe the local one-point strongly convex property. We randomly choose a $\w_0$ and fix it. Then with the initialization $\w_0$, we train ResNet20 on a subset of cifar10 with the size $|S| \in \{10n\%,20n\%,\ldots,100n\%\}$~($n = 50000$). We run momentum SGD to estimate the local minimum around $\w_0$. For each $|S|$, we repeat experiments 10 times. The result is in Figure \ref{exp:ass1}. We can find that $h_S(\w_0) - \mathop{\min}_{\w\in B(\w_0,r)}h_S(\w)$ is almost proportional to $|S|$. As long as $\nabla h_S(\w_{h_S,B(\w_0,r)}^*) \neq \0$, $\|\w_0 - \w_{h_S,B(\w_0,r)}^*\| = r$ so that Figure \ref{exp:ass1} implies the local one-point strongly convex property. Hence, Definition~\ref{def:local strongly convex} is also reasonable for non-convex functions.

With the above assumption and definition, we can obtain the following theorem.
\begin{theorem}\label{theorem:insignificant}
Let $\w_0\in\RB^d, r>0, S \subseteq [n], S^c = [n]/S, \Omega = B(\w_0,r)$ and $h(\cdot)$ is $\mu$-local one-point  strongly convex at $\w_0$. If
\begin{align*}
  \zeta = \frac{2\sum_{i\in S^c}L_i(\w_0,r)}{\mu|S|\|\w_0 - \w_{h_S,\Omega}^*\|} < 1,
\end{align*}
then $h_{S^c}(\cdot)$ is $\zeta$-insignificant at $\w_0$ w.r.t $(h_S(\cdot),\Omega)$. Specifically, if
\begin{align*}
  \zeta = \frac{2\sum_{i\in S^c}L_i(\w_0,r)}{\mu n\|\w_0 - \w_{h,\Omega}^*\|} < 1,
\end{align*}
then $h_{S^c}(\cdot)$ is $\zeta$-insignificant at $\w_0$ w.r.t $(h(\cdot),\Omega)$.
\end{theorem}
According to the theorem, to make $h_{S^c}(\cdot)$ insignificant, we should set $S$ with large Lipschitz constants as far as possible.

\section{ADASS}
In this section, we give the following notations: we denote $f_S(\w) = \frac{1}{|S|}\sum_{i\in S}f_i(\w), S\subseteq [n]=\{1,2,\ldots,n\}$, and for any function $\phi(\cdot)$, we denote $$M_{\gamma,r}(\w;\phi) = \min_{\w'\in B(\w,r)}\phi(\w) + \frac{1}{2\gamma}\|\w' - \w\|^2.$$
The $M_\gamma(\w;\phi)$ is the well-known Moreau envelope. It has been used for the convergence analysis of non-smooth functions~\citep{chen2018universal,DBLP:journals/siamjo/DavisD19}. In this paper, we use it to evaluate the convergence.

According to the analysis in previous sections, we should pay more attention to those loss function with large local Lipschitz constants. Hence, different training subsets need to be adaptively selected from the full training set for different training states. Because sample selection of our method is adaptive to different training states, we name our method \underline{ada}ptive \underline{s}ample \underline{s}lection~(ADASS). ADASS is presented in Algorithm \ref{alg:adass}.

\begin{algorithm}[t]
\caption{Adaptive Sample Selection~(ADASS)}
\label{alg:adass}
\begin{algorithmic}[1]
\STATE Initialization: $r>0, \gamma>0, \alpha\in(0,1],  S_0 = [n], \w_0$.
\FOR{$t=0,1,\ldots,T$}
\STATE If $mod(t,p)=0$, calculate loss values $\{f_i(\w_t) \}$ and select samples $S_t$ such that
\vspace{-0.3cm}
       \begin{align}\label{eq:LASS_selection}
       \sum_{i\in S_t}|f_i(\w_{t}) - f_i(\w_{t-p})| \geq \alpha\sum_{i=1}^n|f_i(\w_{t}) - f_i(\w_{t-p})|;
       \end{align}
\vspace{-0.3cm}
\STATE Else $S_t = S_{t-1}$;
\STATE Let $\hat{F}_t^\gamma(\w) = |S_t|(f_{S_t}(\w) + \frac{1}{2\gamma}\|\w - \w_t\|^2)$;
\STATE $\w_{t+1} = O(\hat{F}_t^\gamma, \w_{t}, K_t, \eta_t, B(\tilde{\w},r))$;
\ENDFOR
\end{algorithmic}
\end{algorithm}

Since it is difficult to obtain the exact local Lipchitz constant, we use $\frac{|f_i(\w_t) - f_i(\w_{t-1})|}{\|\w_t - \w_{t-1}\|}$ as its approximation. Although it is a rough approximation, it is enough to guarantee the insignificant property. In (\ref{eq:LASS_selection}), obviously, if $\alpha \rightarrow 1$, which means $S_t$ is almost the full training set $[n]$, the corresponding $\zeta$ is close to $0$. Thus, if we need to select samples, we choose those with large $|f_i(\w_t) - f_i(\w_{t-1})|$.

After sample selection, ADASS is going to optimize $\hat{F}_t^\gamma(\w)$ which is defined as:
\begin{align*}
	\hat{F}_t^\gamma(\w) = |S_t|(f_{S_t}(\w) + \frac{1}{2\gamma}\|\w - \w_{t}\|^2)
\end{align*}
With weakly convex assumption and suitable $\gamma$, $\hat{F}_t^\gamma(\w)$ will be strongly convex. Hence, it is easy to guarantee the insignificant property according to Lemma \ref{lemma:strongconvex} and Theorem \ref{theorem:insignificant}. ADASS is mainly focus on sample selection, it can adopt any existing optimization tools, denoted as $O(\hat{F}_t^\gamma, \w_{t}, K_t, \eta_t, B(\w_{t},r))$, $\w_t$ is the initialization, $K_t$ is the number of iterations, $\eta_t$ is the constant step size, $B(\w_{t},r)$ is the optimization domain. We assume that $\w_{t+1} = O(\hat{F}_t^\gamma, \w_t, K_t, \eta_t, B(\w_{t},r))$ satisfies that
\begin{align}
	\EB \hat{F}_t^\gamma(\w_{t+1}) - \EB \hat{F}_t^\gamma(\w) \leq \theta_{1,t}\|\w_t - \w\|^2 + \theta_{2,t}, \forall \w \in B(\w_{t},r).
\end{align}

For convenience, let
\begin{align*}
	\bar{F}_t^\gamma(\w) = (n-|S_t|)(f_{S_t^c}(\w) + \frac{1}{2\gamma}\|\w - \w_{t}\|^2), ~~F_t^\gamma(\w) = n(f_{[n]}(\w) + \frac{1}{2\gamma}\|\w - \w_{t}\|^2),
\end{align*}
then we have the following convergence result:
\begin{theorem}\label{theorem:adass_converge1}
Assume that $f_i(\cdot)$ is $c$-weakly convex, $\bar{F}_t^{\gamma}(\cdot)$ is $\zeta_{1,t}$-insignificant at $\w_{t}$, $\zeta_{2,t}$-insignificant at $\w_{t+1}$  w.r.t $(\hat{F}_t^{\gamma}(\cdot), B(\w_{t},r))$, and $\zeta_{1,t} \leq 1, \zeta_{2,t}\leq \zeta_2$, $\EB F(\w_t) - F(\w^*) \leq \delta$. By setting $\gamma = 1/2c, \theta_{1,t} \leq \frac{n}{24\gamma (1+\zeta_2)|S_t|}, \theta_{2,t} \leq \theta_2/t$, we have
\begin{align*}
\frac{2}{T(T+1)}\sum_{t=1}^Tt\EB\|\nabla M_{\gamma,r}(\w_t;f_{S_t})\|^2 \leq \frac{16\delta}{\gamma(T+1)} + \frac{48\theta_2}{\gamma T(T+1)}\sum_{t=1}^T\frac{(1+\zeta_{2,t})|S_t|}{n}.
\end{align*}
\end{theorem}

In Theorem \ref{theorem:adass_converge1}, we denote
\begin{align}\label{eq:rho}
\rho_T = \frac{1}{T}\sum_{t=1}^T\frac{(1+\zeta_{2,t})|S_t|}{n}.
\end{align}

If we set $S_t = [n], \forall t$, then $\rho_T$ = 1 and ADASS degenerates to normal optimization methods. If $\rho_T < 1$, then ADASS gets faster convergence rate. We will show that $\rho_T < 1$ in the empirical results. In the above theorem, we proof the convergence of $f_{S_t}(\w_t)$. According to Property \ref{pro:2}, it is enough to guarantee the convergence of $f_{[n]}(\w_t)$. In the next theorem, we also get the convergence of $f_{[n]}(\w_t)$ directly with certain assumptions as follow:
\begin{theorem}\label{theorem:adass_converge2}
Assume that $f_i(\cdot)$ is $c$-weakly convex, $\bar{F}_t^{\gamma}(\cdot)$ is $\zeta_{t}$-insignificant at $\w_{t+1}$  w.r.t $(F_t^{\gamma}(\cdot), B(\w_{t},r))$, and $\zeta_{t}\leq \zeta<1$, $\EB F(\w_t) - F(\w^*) \leq \delta$. By setting $\gamma = 1/2c, \theta_{1,t} \leq \frac{n(1-\zeta)}{24\gamma |S_t|}, \theta_{2,t} \leq \theta_2/t$, we have
\begin{align*}
	 \frac{2}{T(T+1)}\sum_{t=1}^T t\EB\|\nabla M_{\gamma,r}(\w_t;f_{[n]})\|^2 \leq \frac{16\delta}{\gamma(T+1)} + \frac{48\theta_2}{\gamma T(T+1)}\sum_{t=1}^T\frac{|S_t|}{n(1-\zeta_t)}.
\end{align*}

\end{theorem}

\section{Experiments}
We conduct experiments to evaluate ADASS with the optimization methods momentum SGD in Algorithm~\ref{alg:adass}. We consider two datasets: CIFAR10, CIFAR100 and two models: AlexNet, ResNet20. We set three different values for the $c$ in Algorithm~\ref{alg:adass}: $\alpha=1$, which is equivalent to normal momentum SGD~(MSGD), $\alpha=0.99$, which we denote as ADASS(v1) and $\alpha=0.999$, which we denote as ADASS(v2). All the experiments are conducted on the Pytorch platform with GPU Titan XP.

\begin{figure*}[htb]
\centering
\subfigure{
  \includegraphics[width =3.5cm]{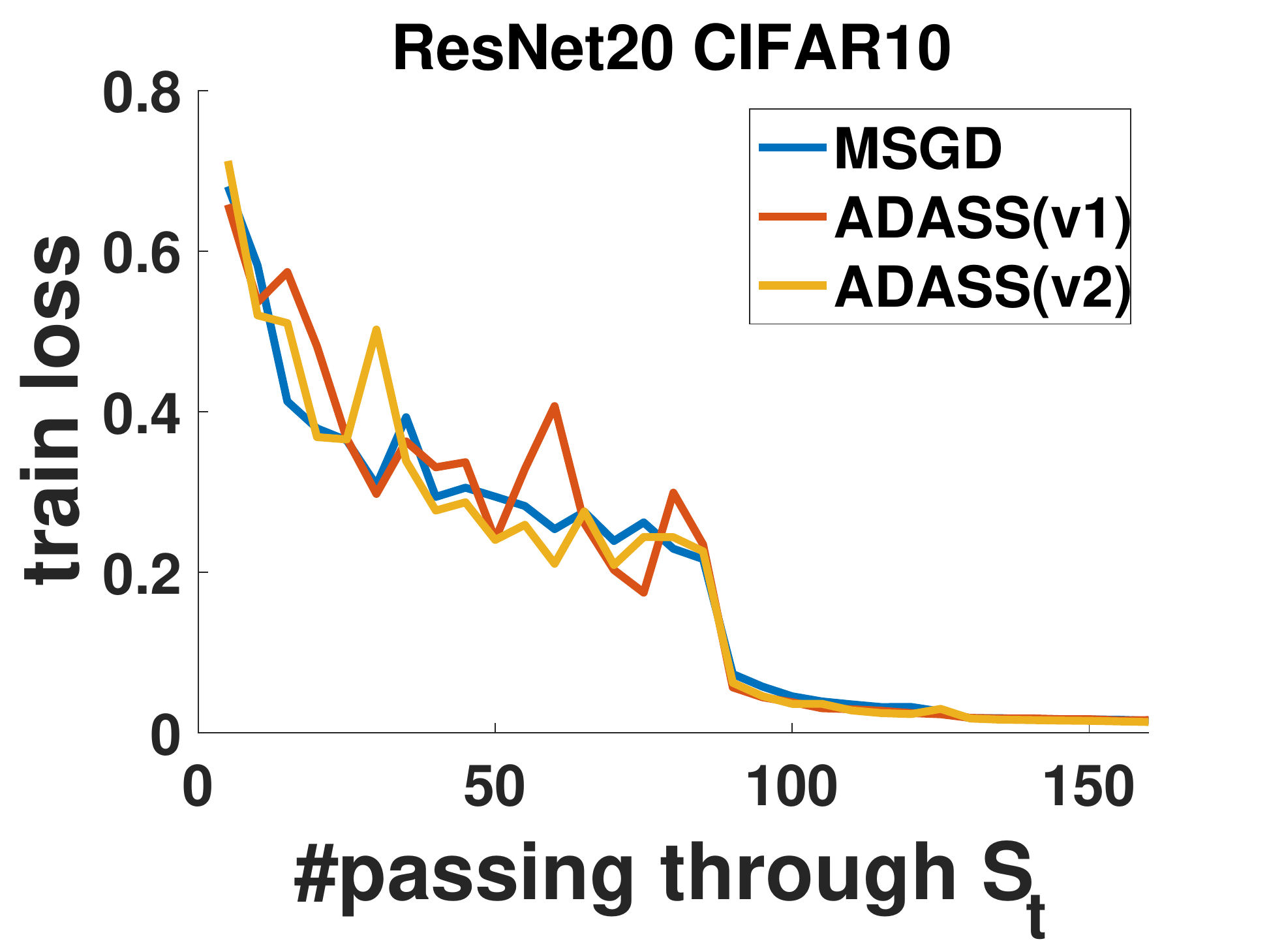}
  \includegraphics[width =3.5cm]{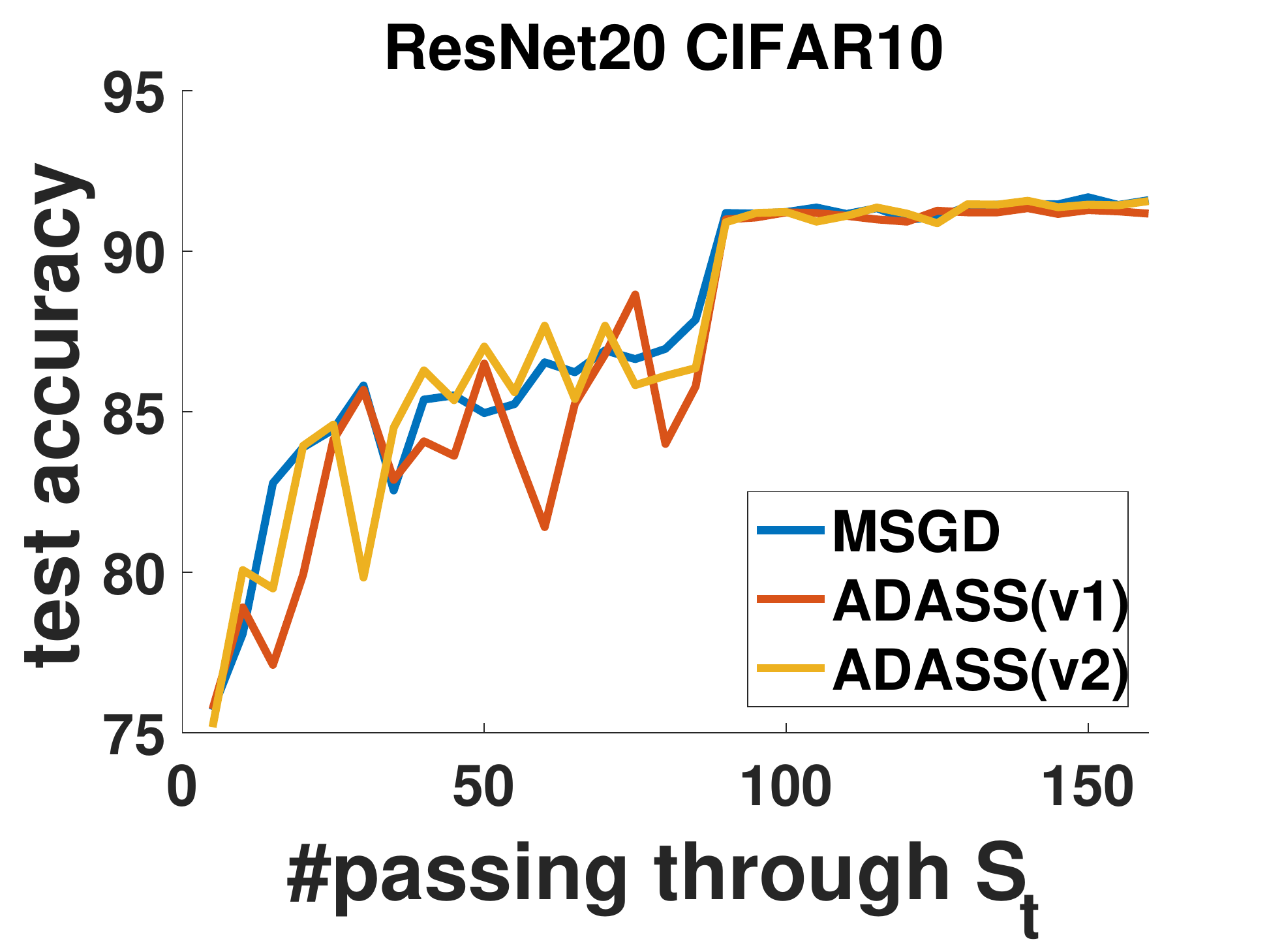}
  \includegraphics[width =3.5cm]{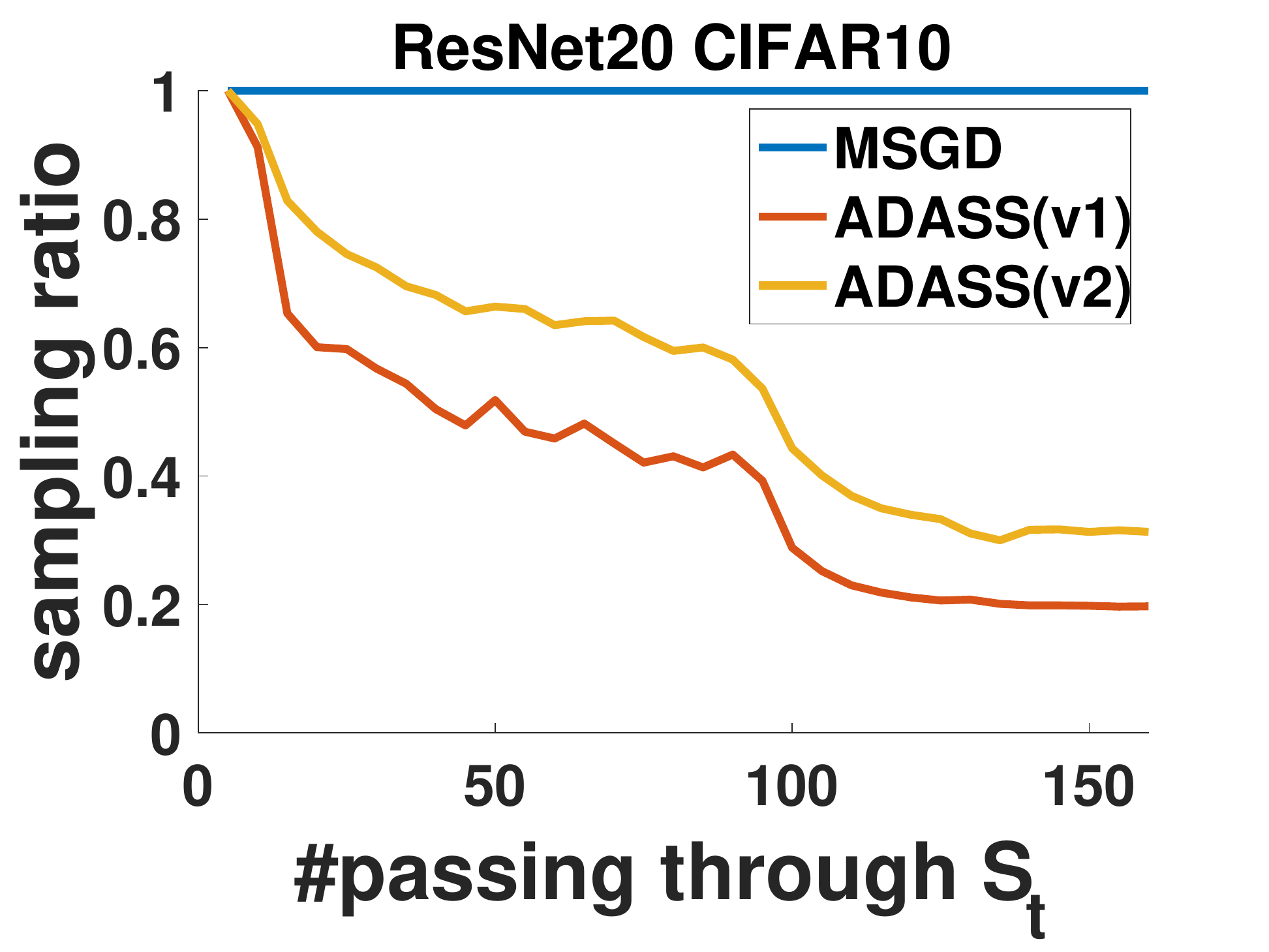}
  \includegraphics[width =3.5cm]{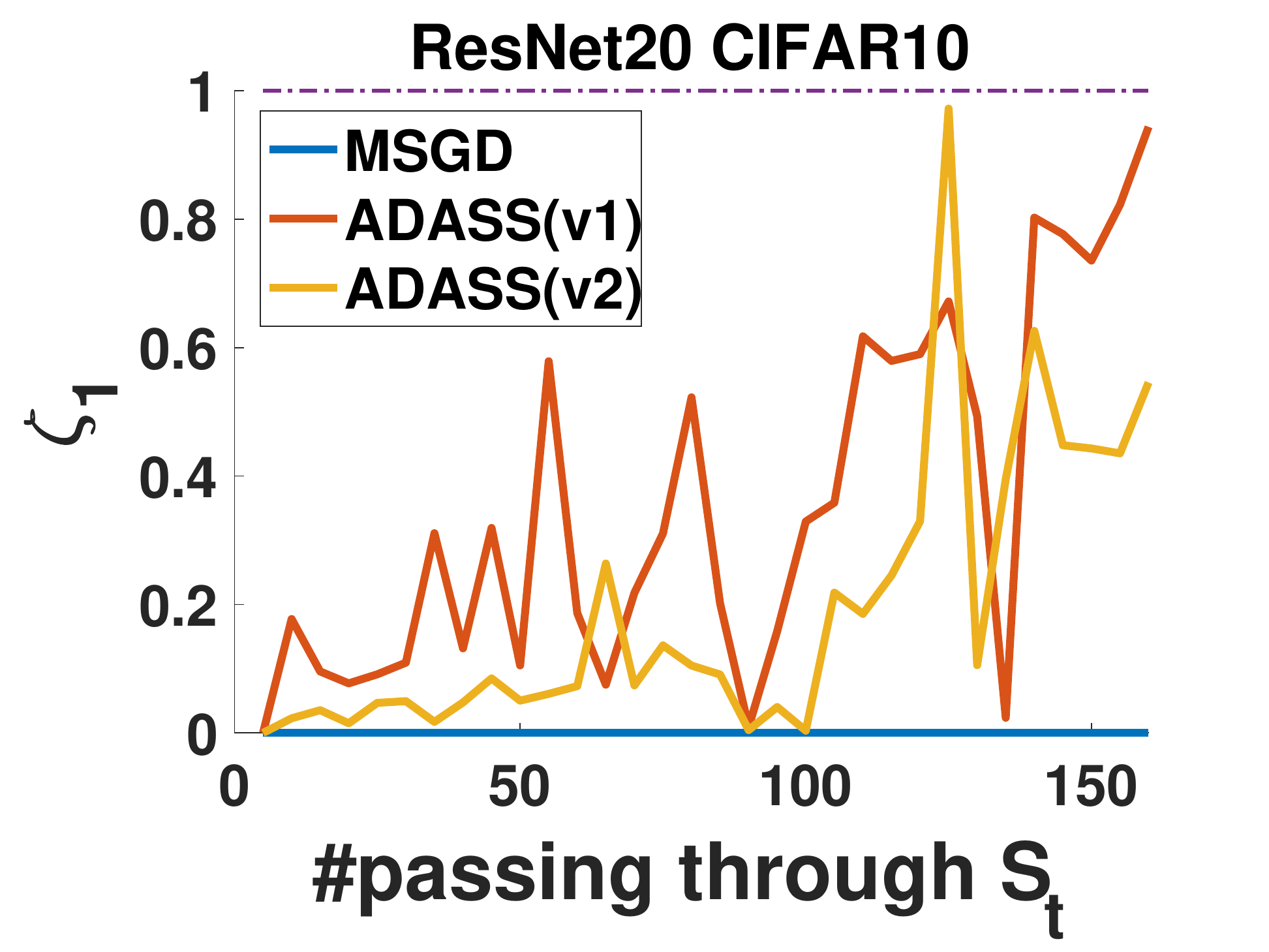}}
\subfigure{
  \includegraphics[width =3.5cm]{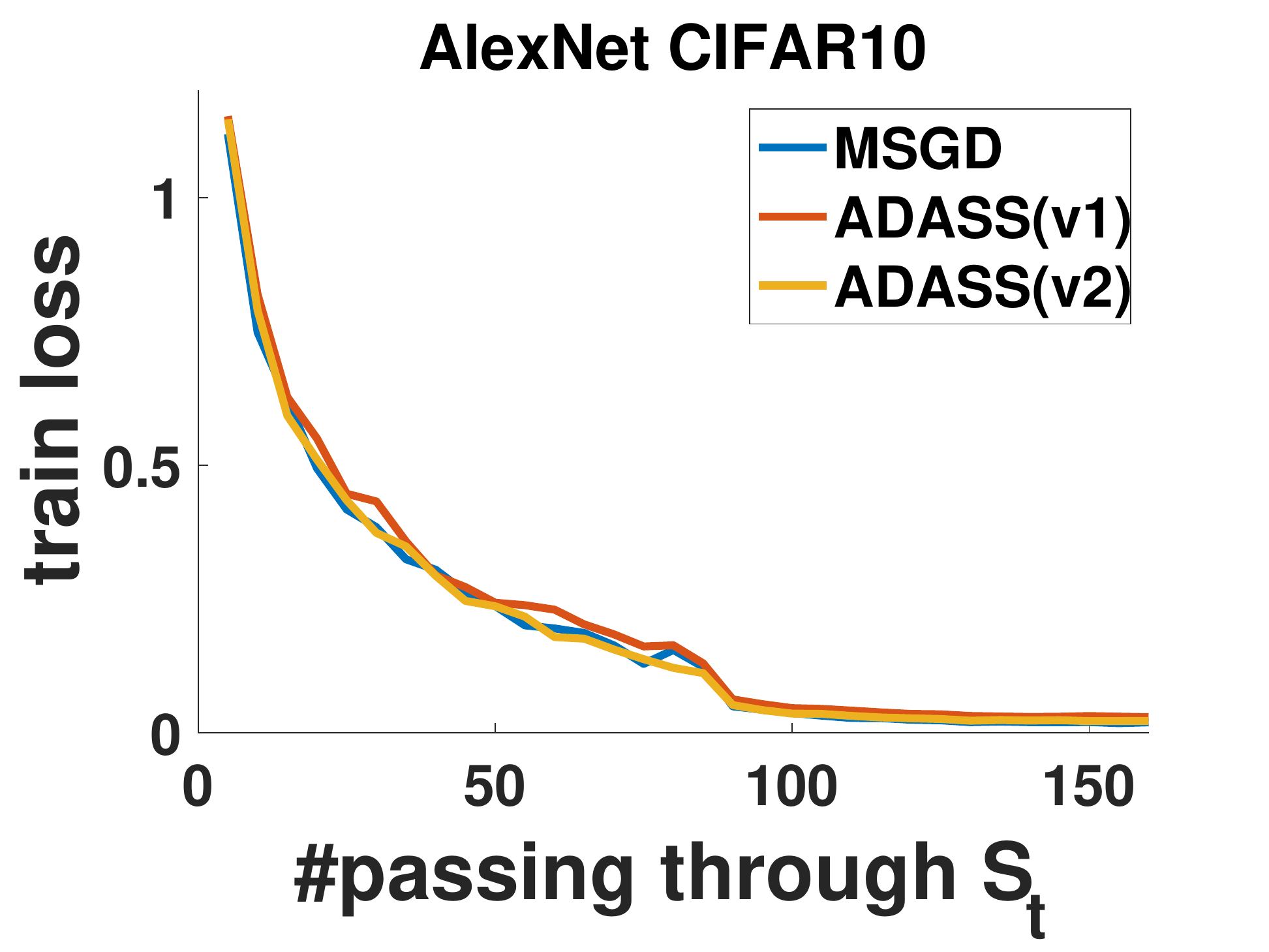}
  \includegraphics[width =3.5cm]{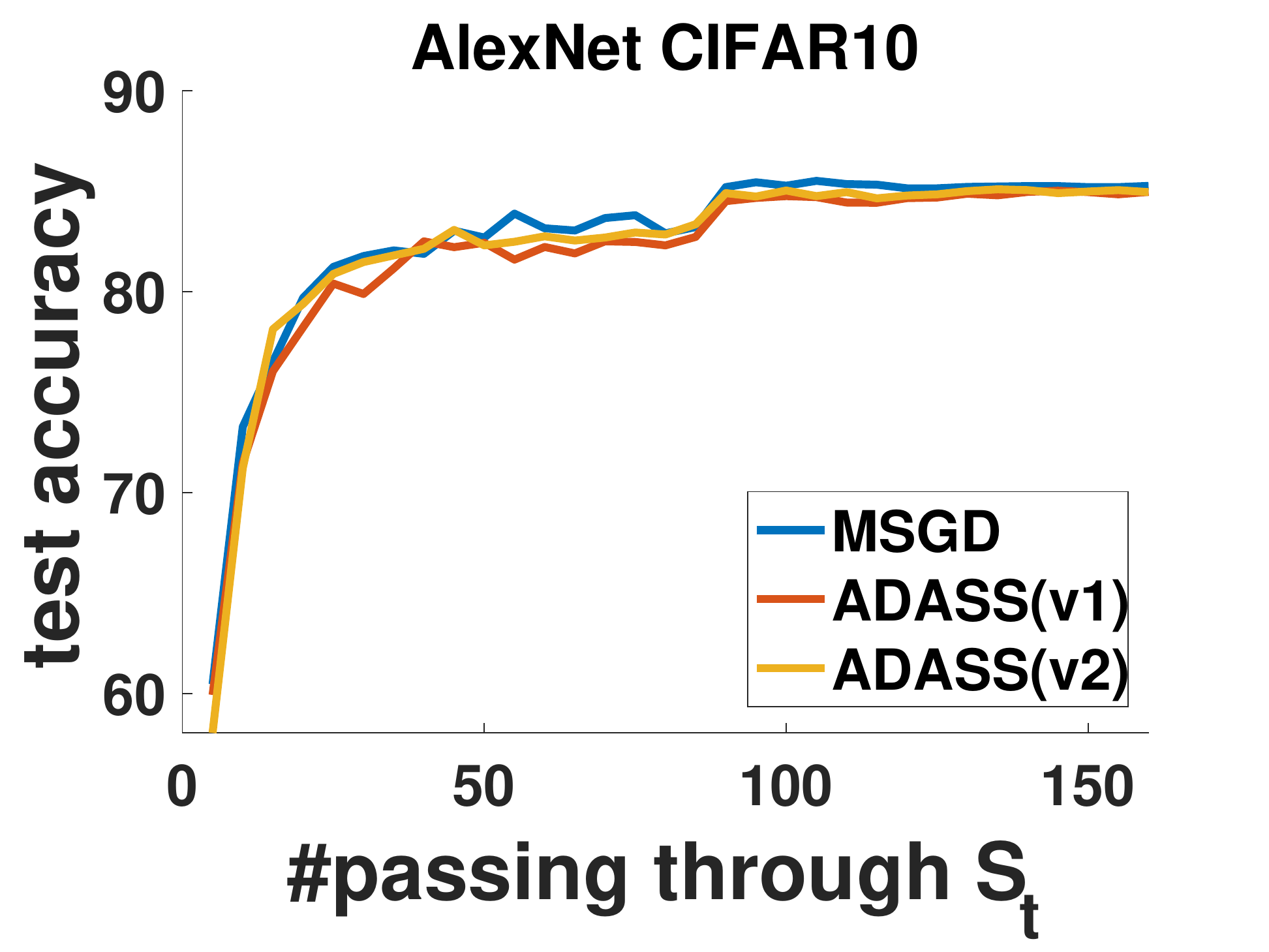}
  \includegraphics[width =3.5cm]{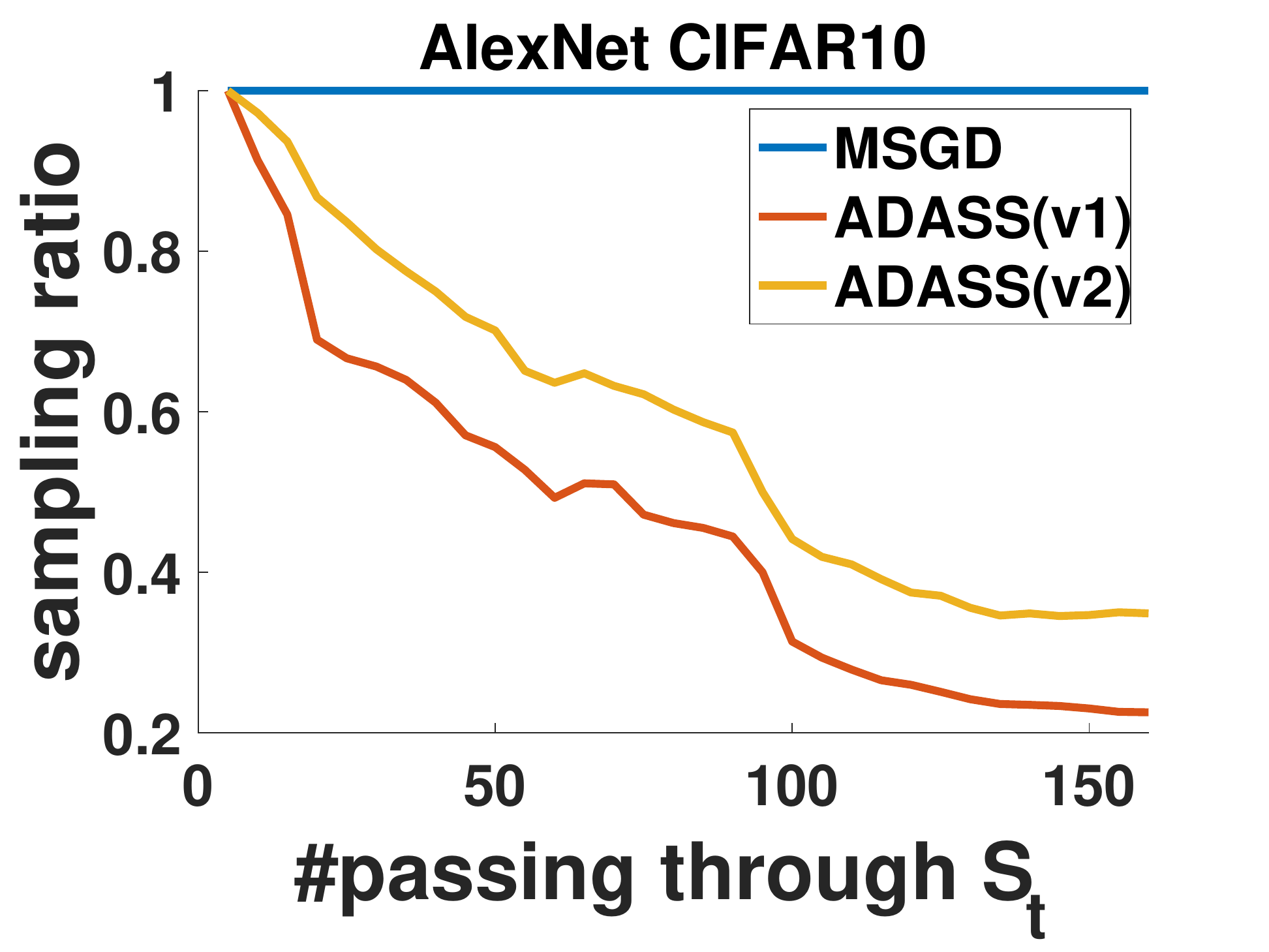}
  \includegraphics[width =3.5cm]{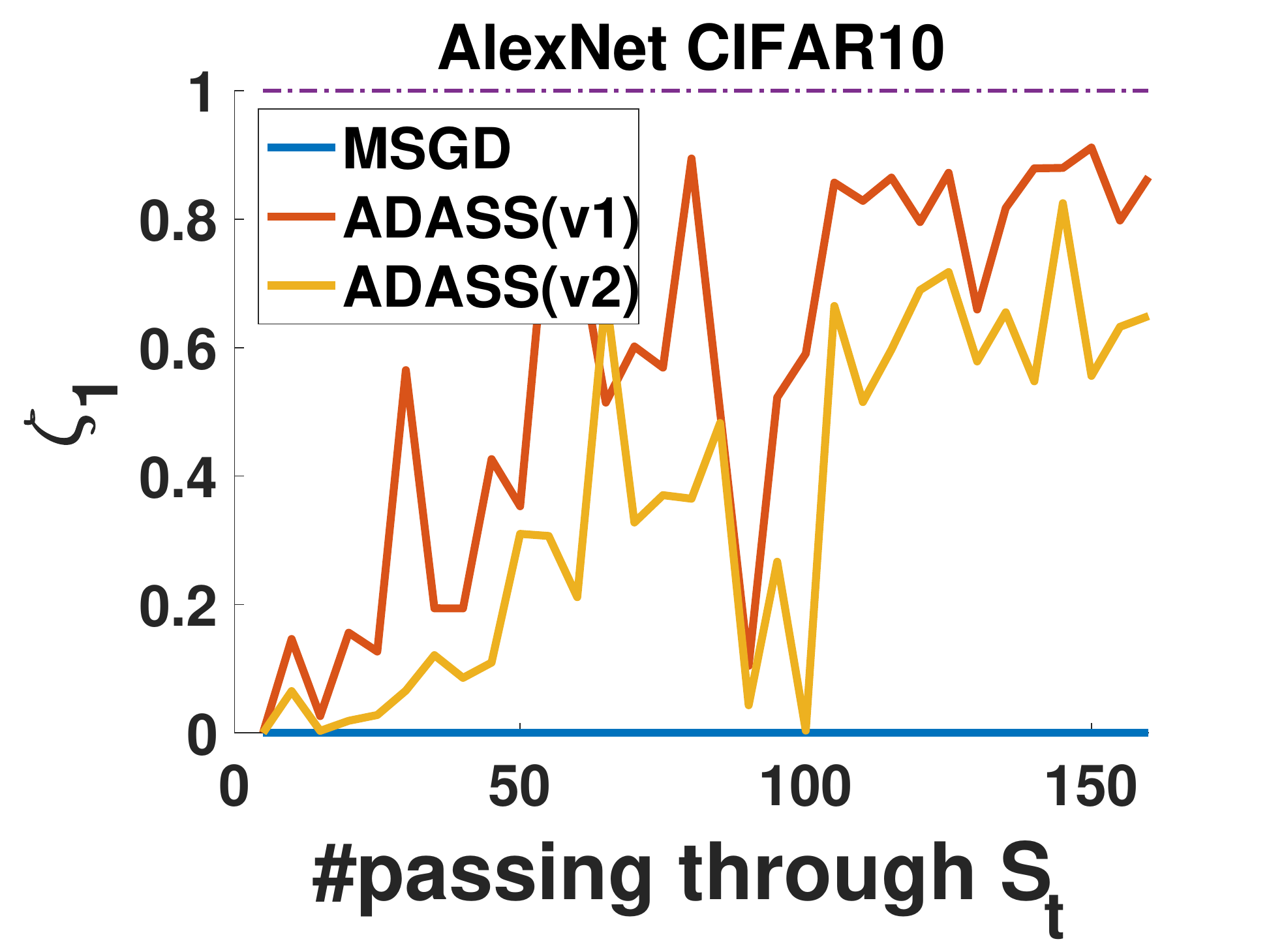}}
\subfigure{
  \includegraphics[width =3.5cm]{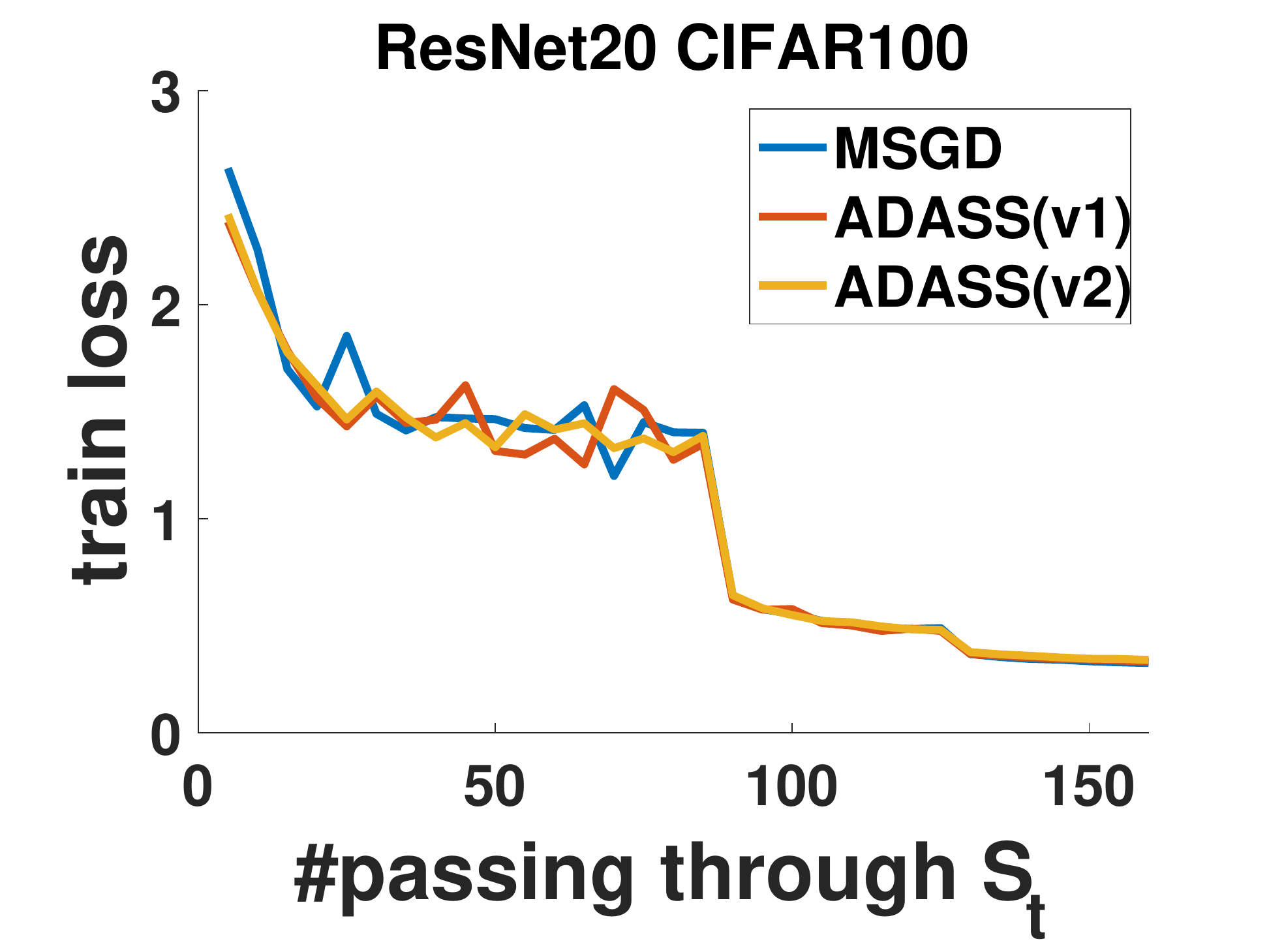}
  \includegraphics[width =3.5cm]{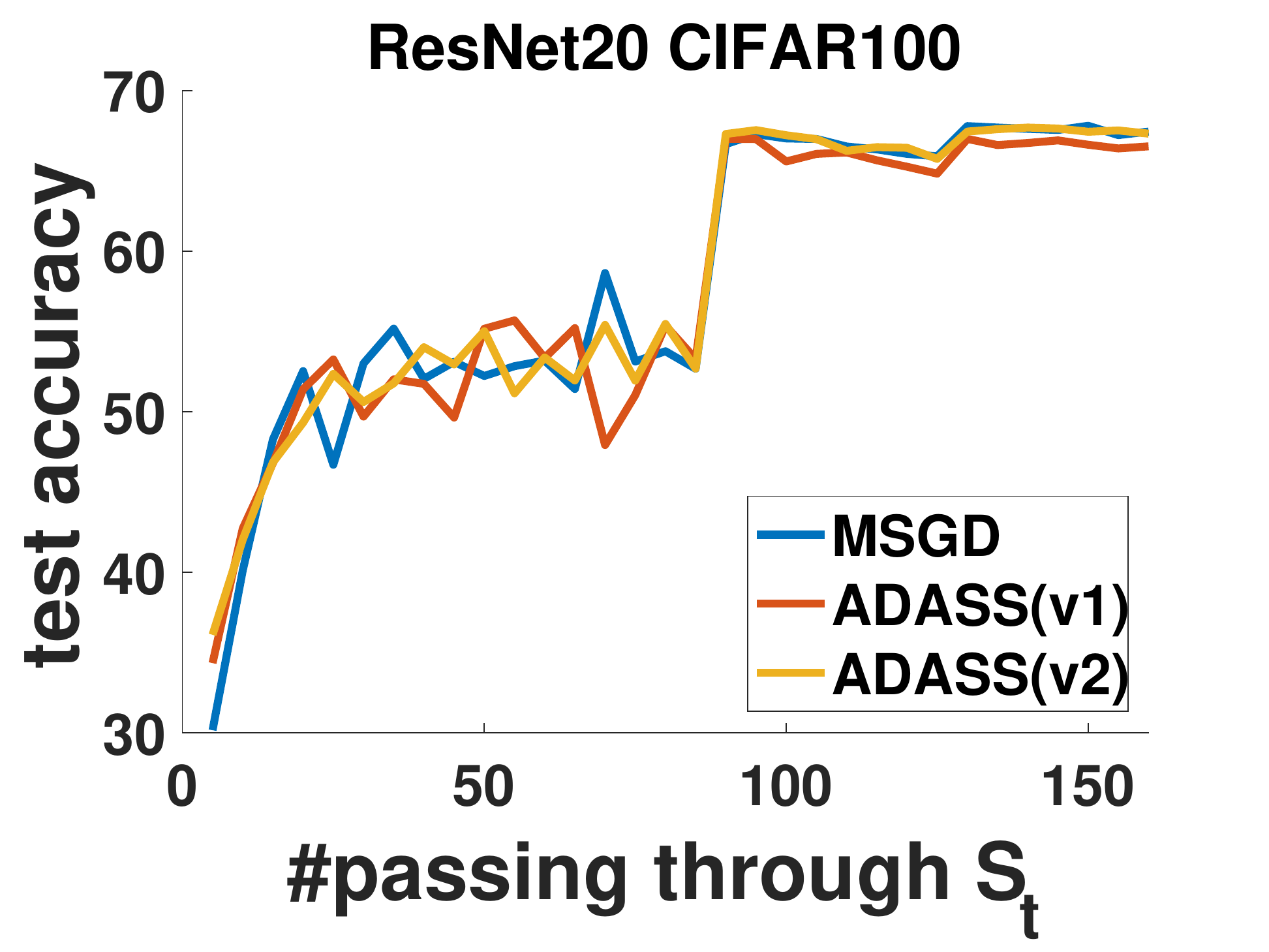}
  \includegraphics[width =3.5cm]{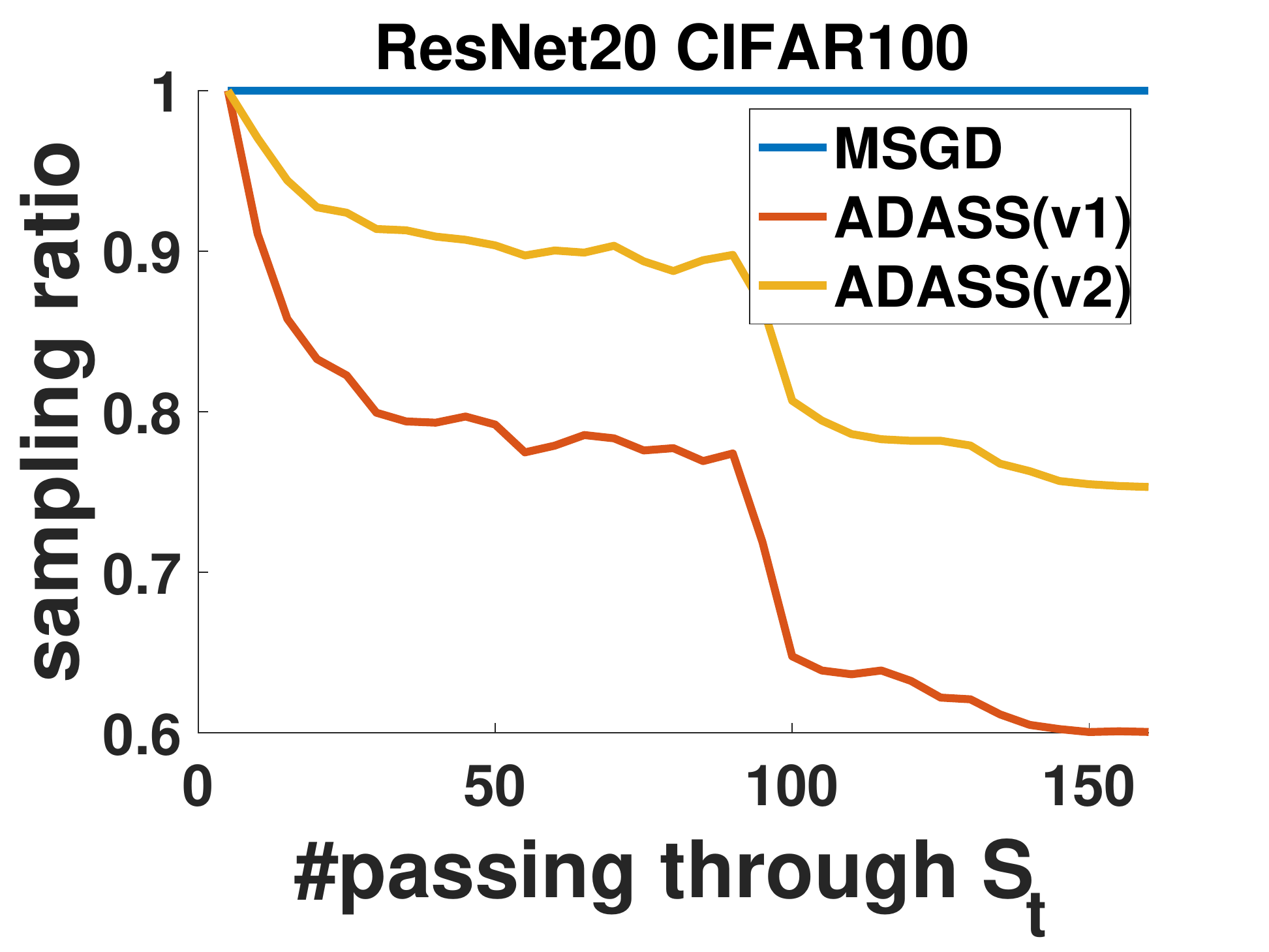}
  \includegraphics[width =3.5cm]{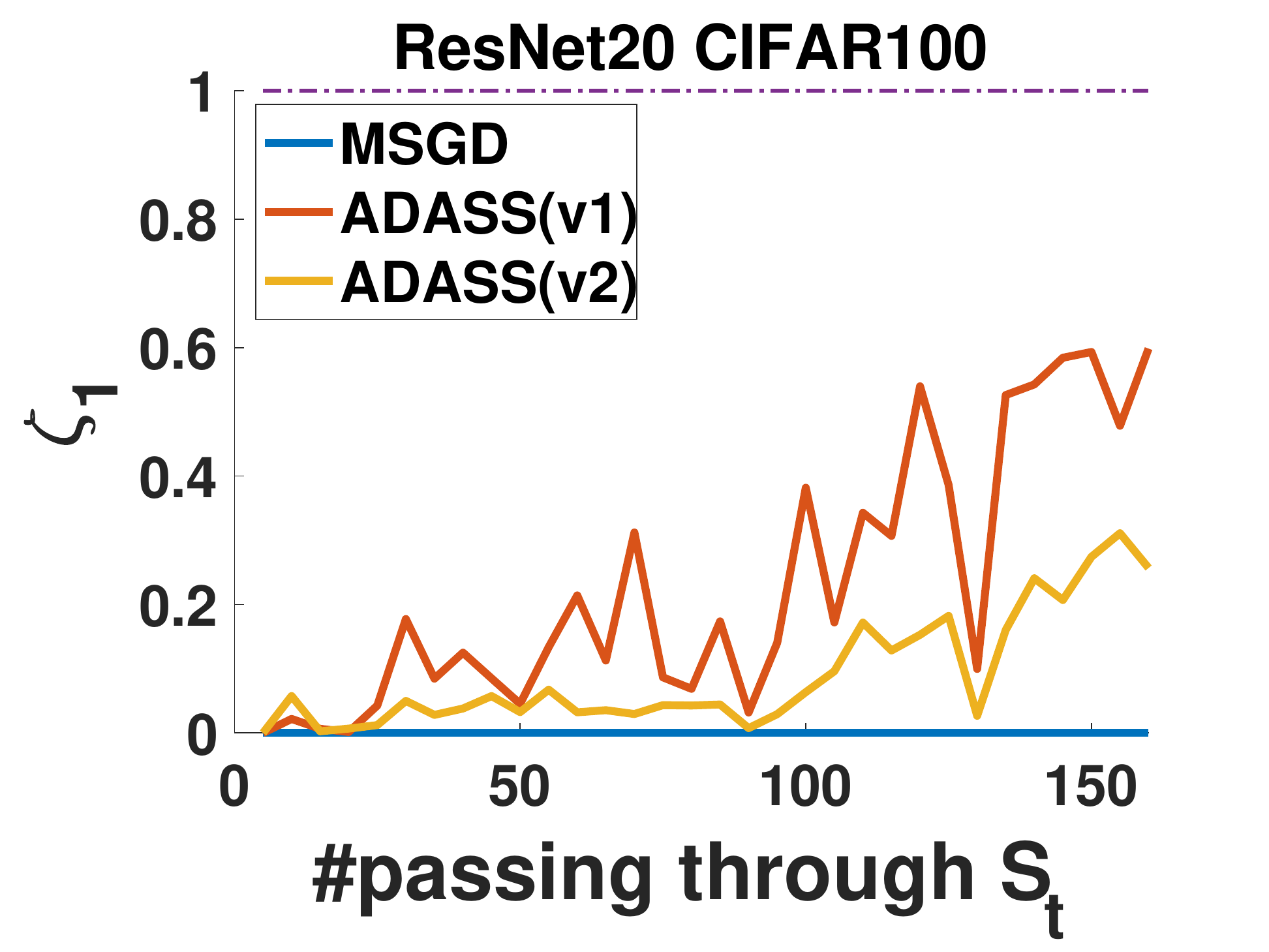}}
\subfigure{
  \includegraphics[width =3.5cm]{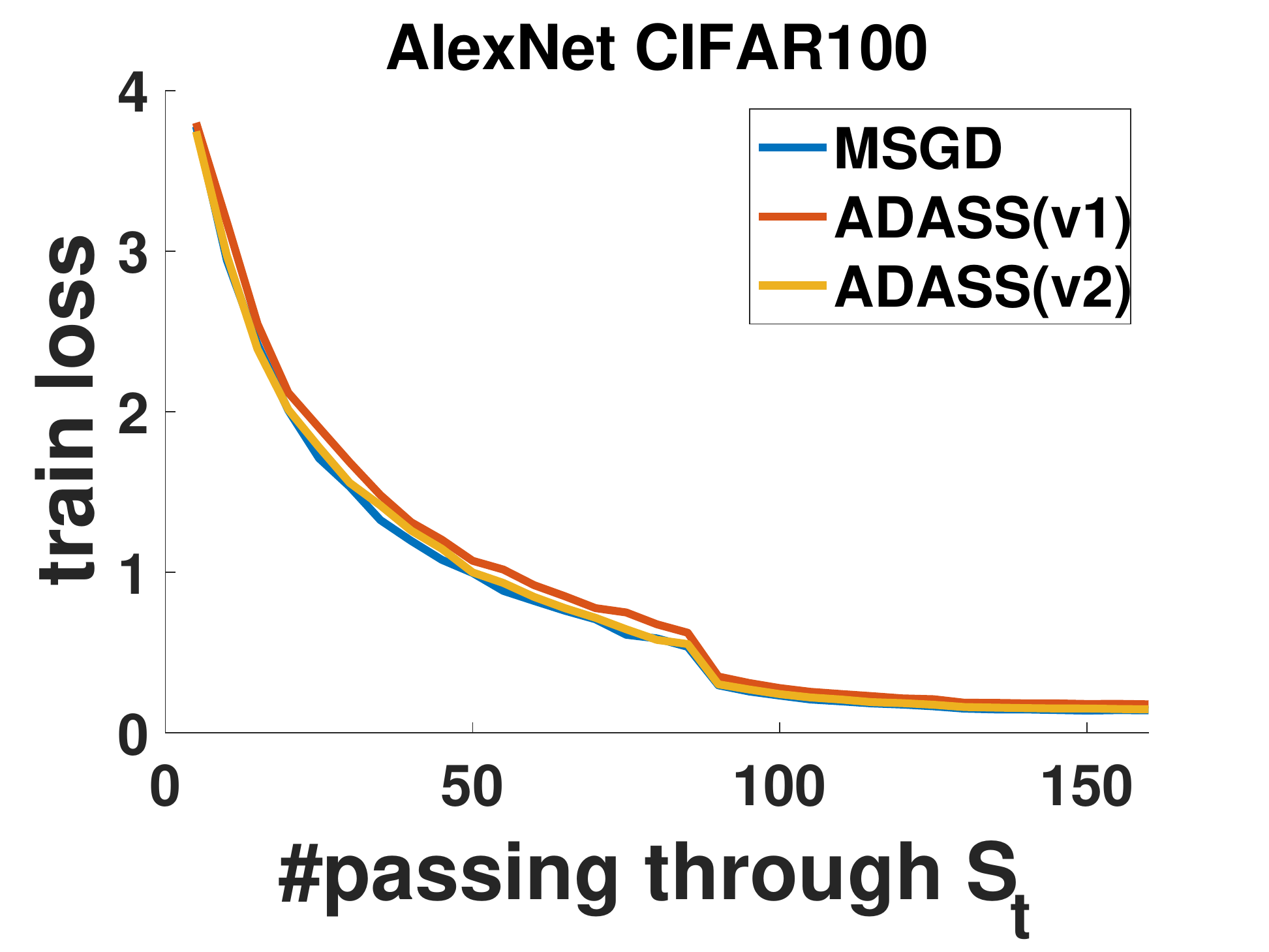}
  \includegraphics[width =3.5cm]{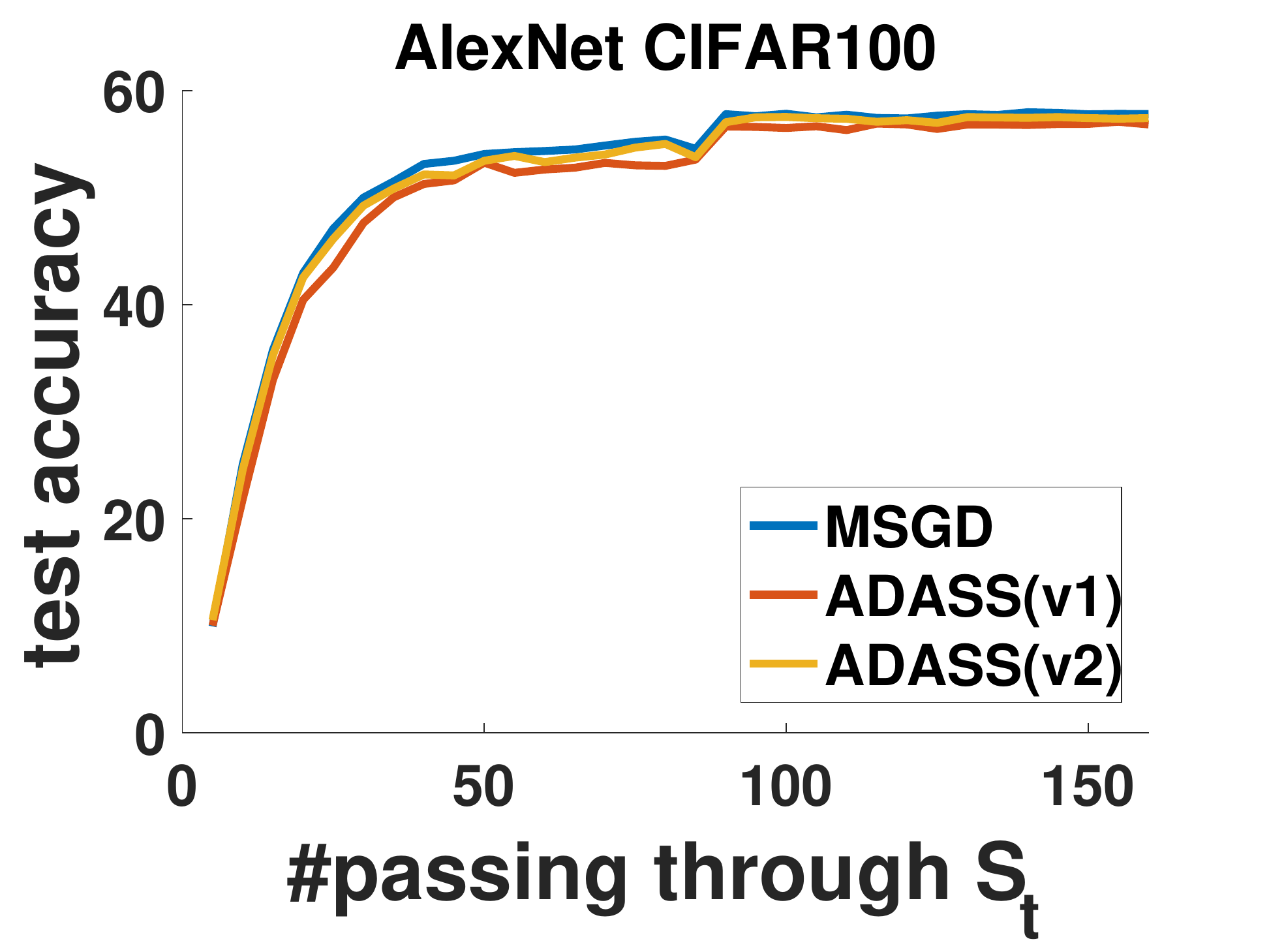}
  \includegraphics[width =3.5cm]{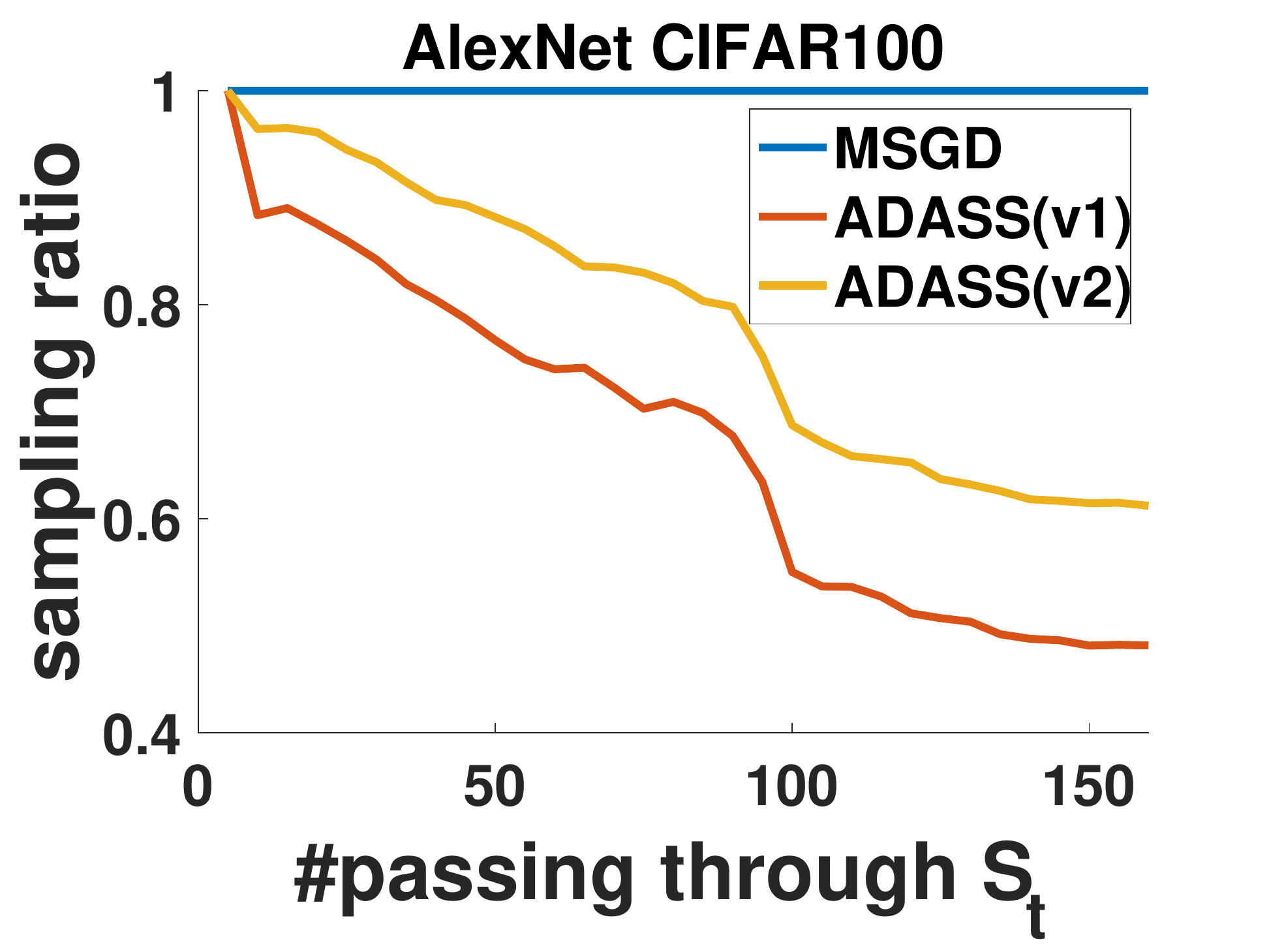}
  \includegraphics[width =3.5cm]{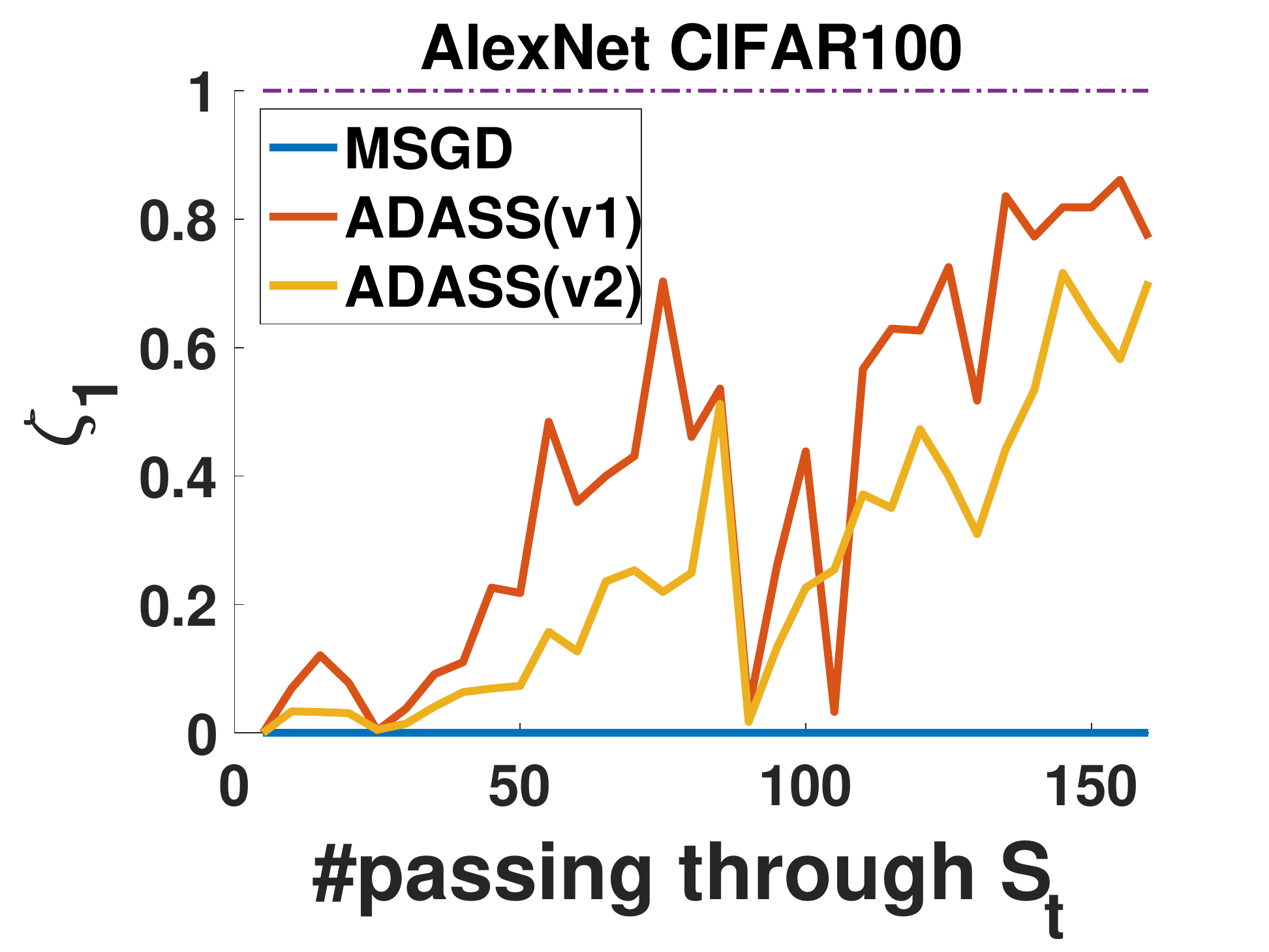}}
  \caption{Train ResNet20/AlexNet on CIFAR10/CIFAR100.}\label{exp:gamma_infty}
\end{figure*}

\begin{figure*}[htb]
\centering
\subfigure{
  \includegraphics[width =3.5cm]{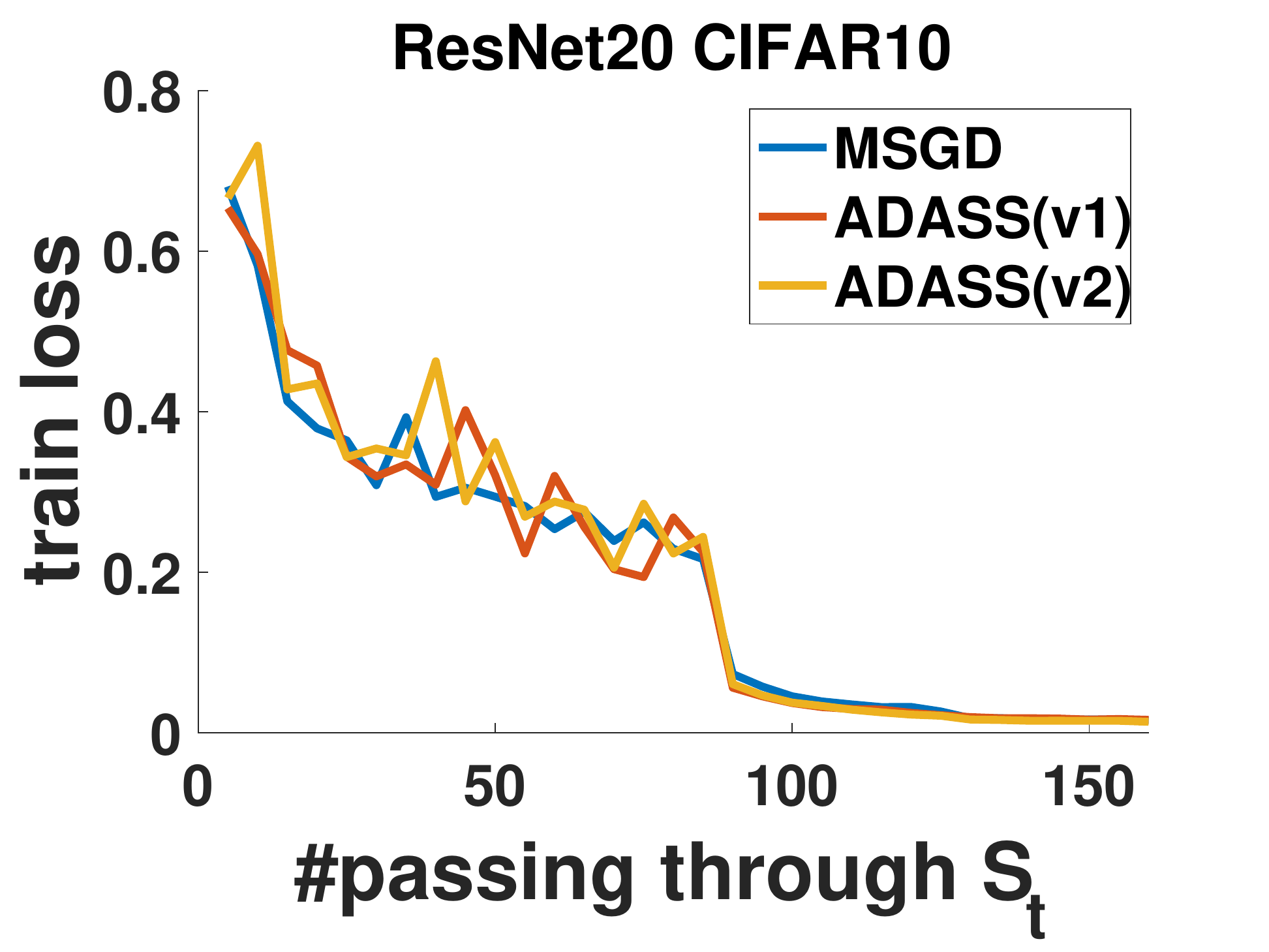}
  \includegraphics[width =3.5cm]{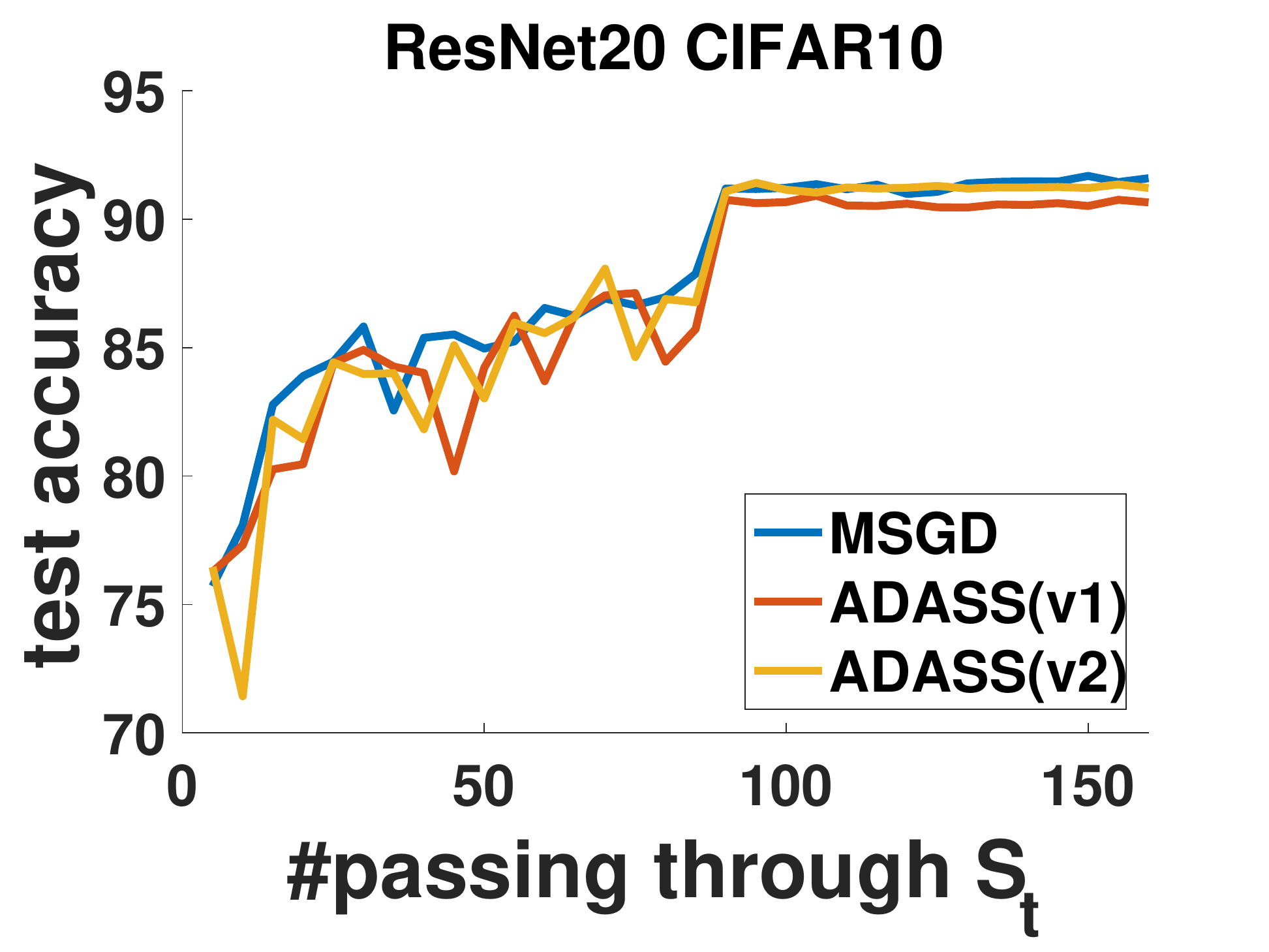}
  \includegraphics[width =3.5cm]{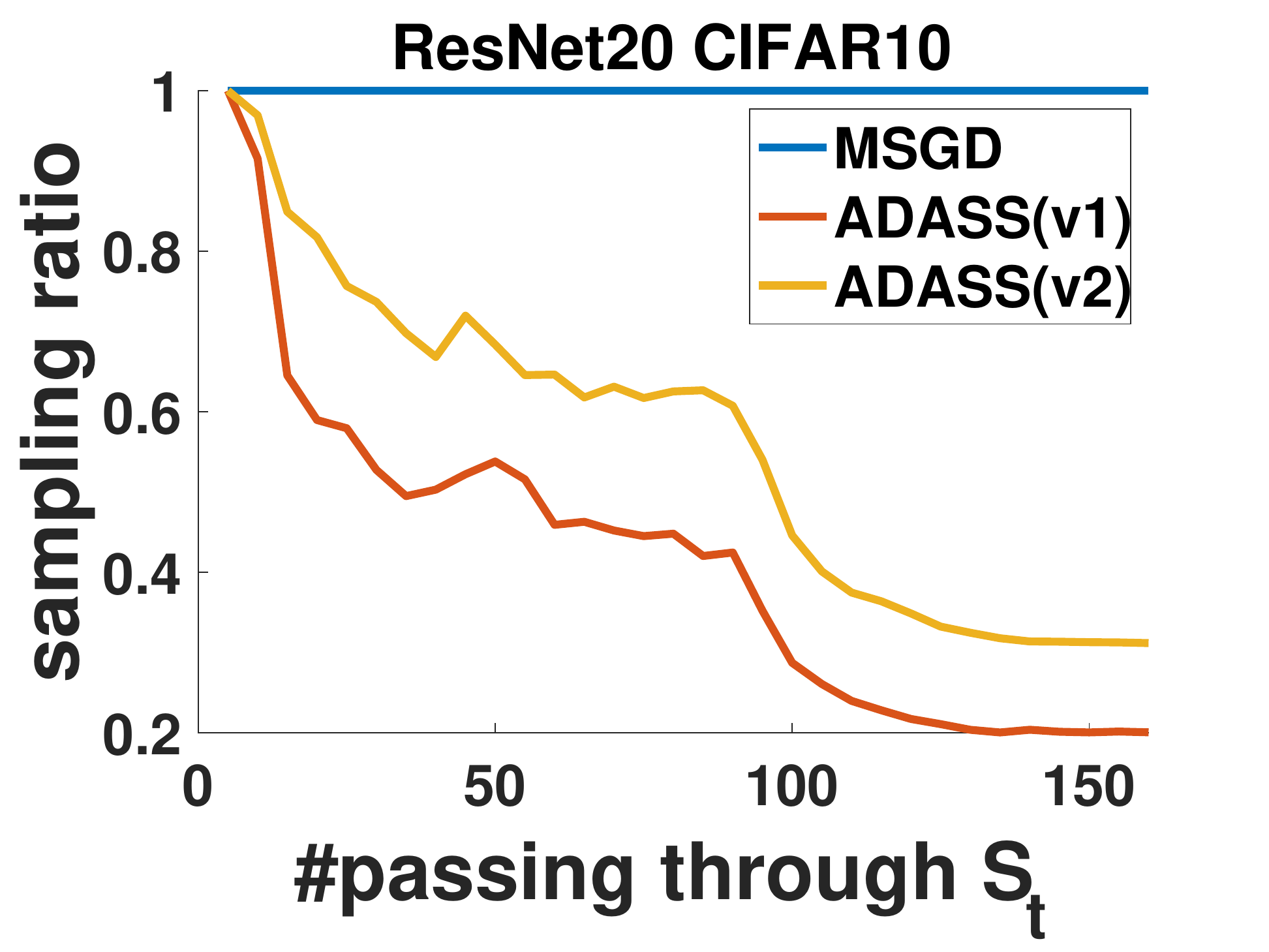}
  \includegraphics[width =3.5cm]{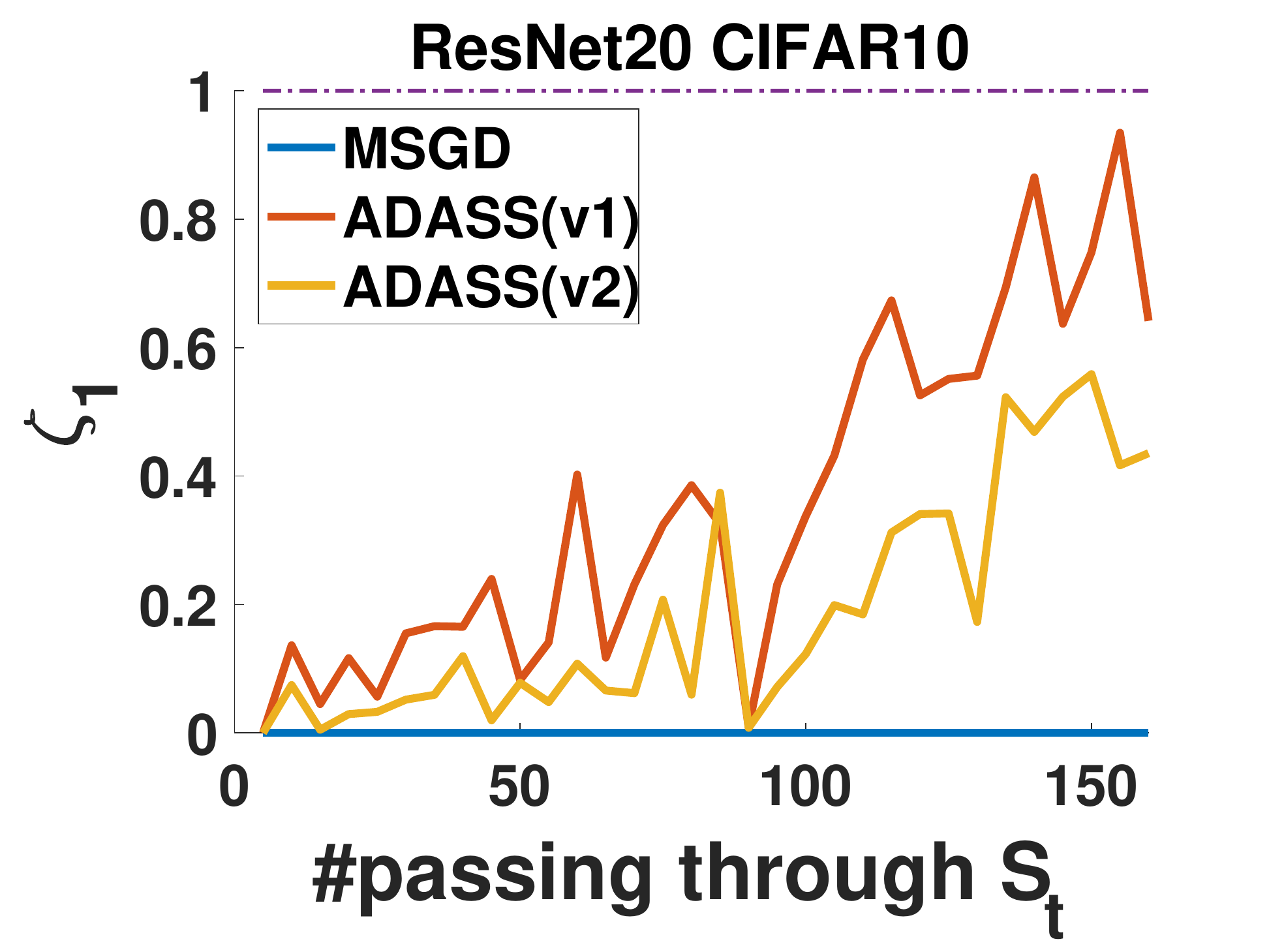}}
\subfigure{
  \includegraphics[width =3.5cm]{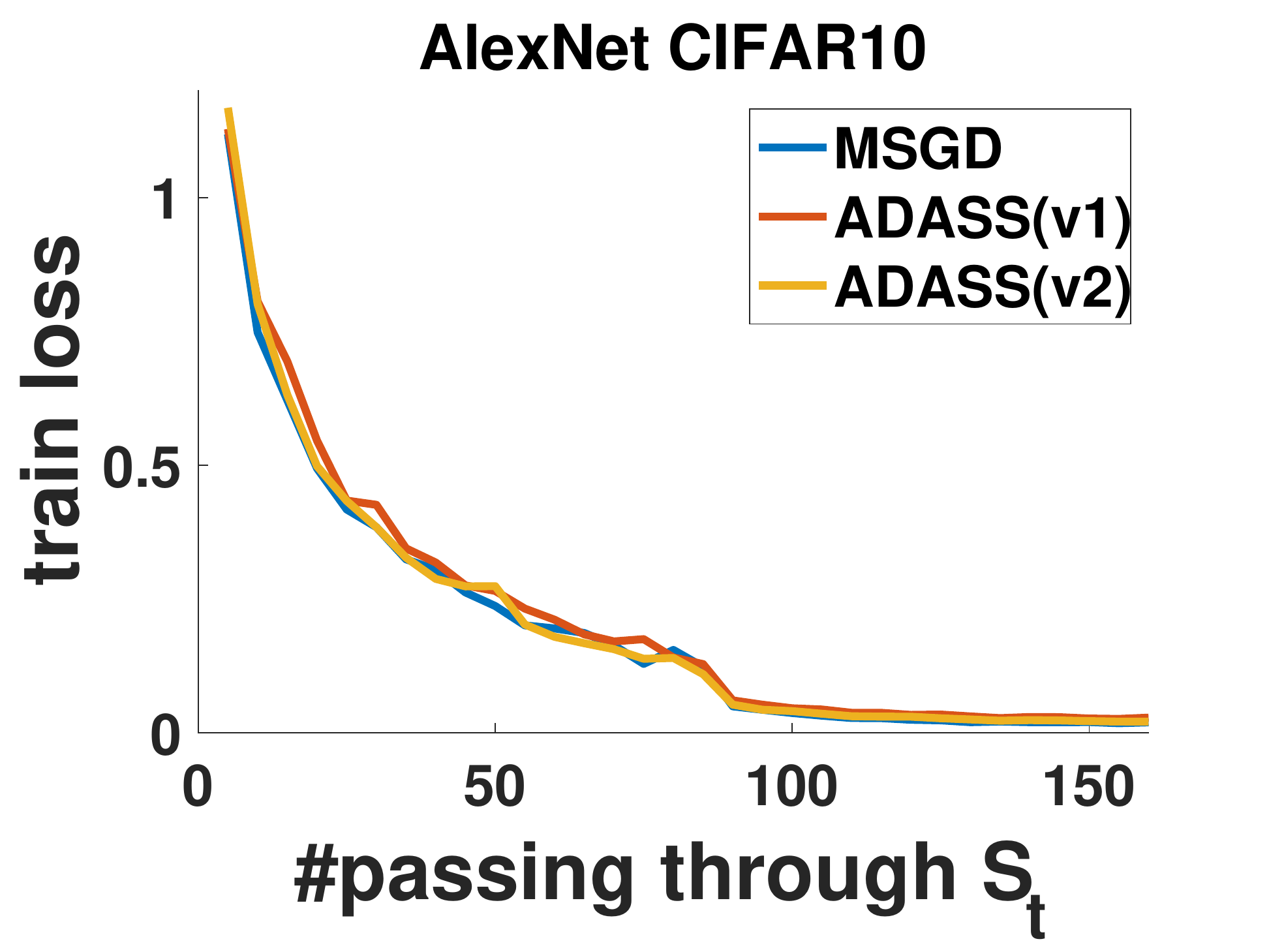}
  \includegraphics[width =3.5cm]{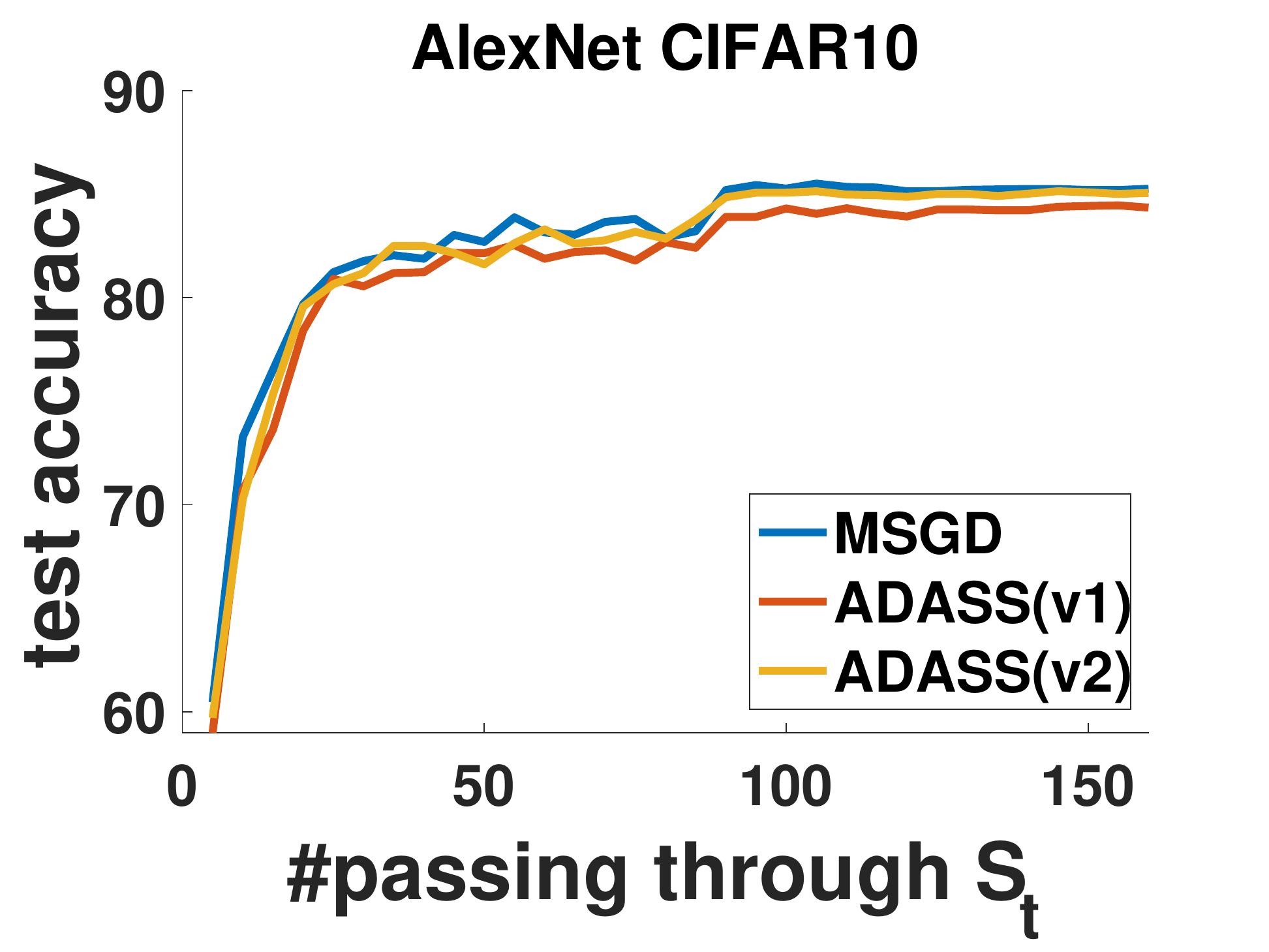}
  \includegraphics[width =3.5cm]{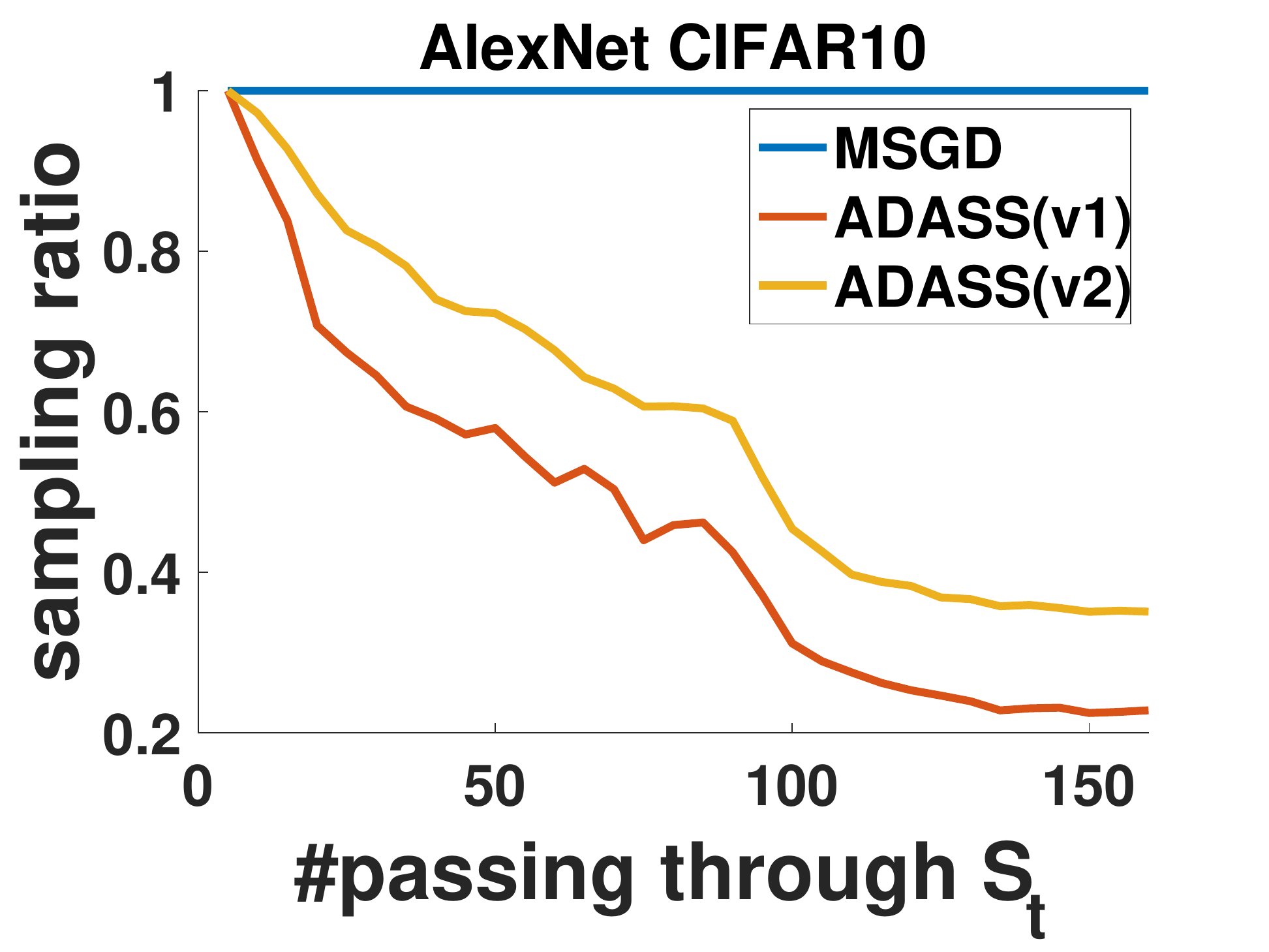}
  \includegraphics[width =3.5cm]{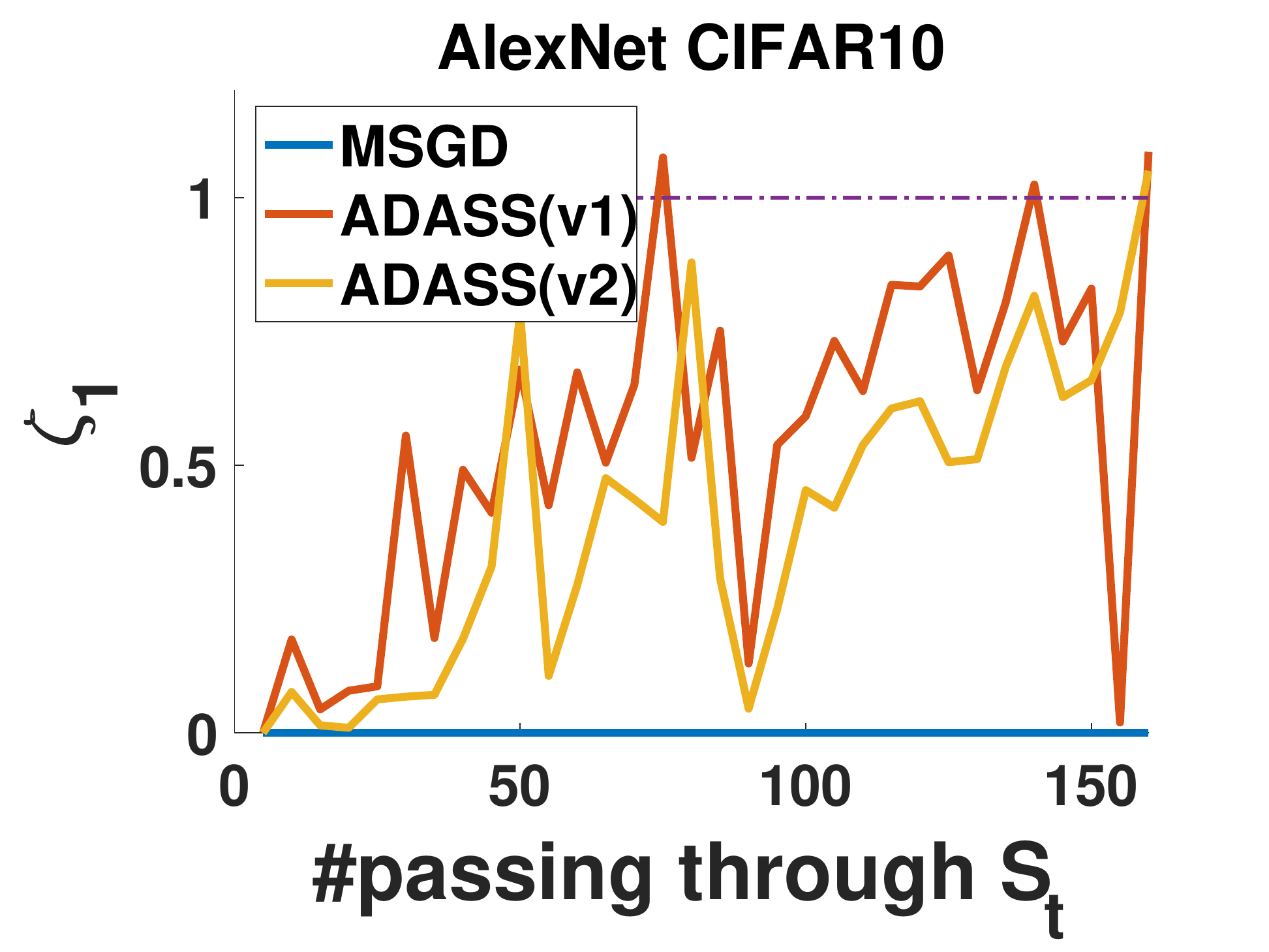}}
\subfigure{
  \includegraphics[width =3.5cm]{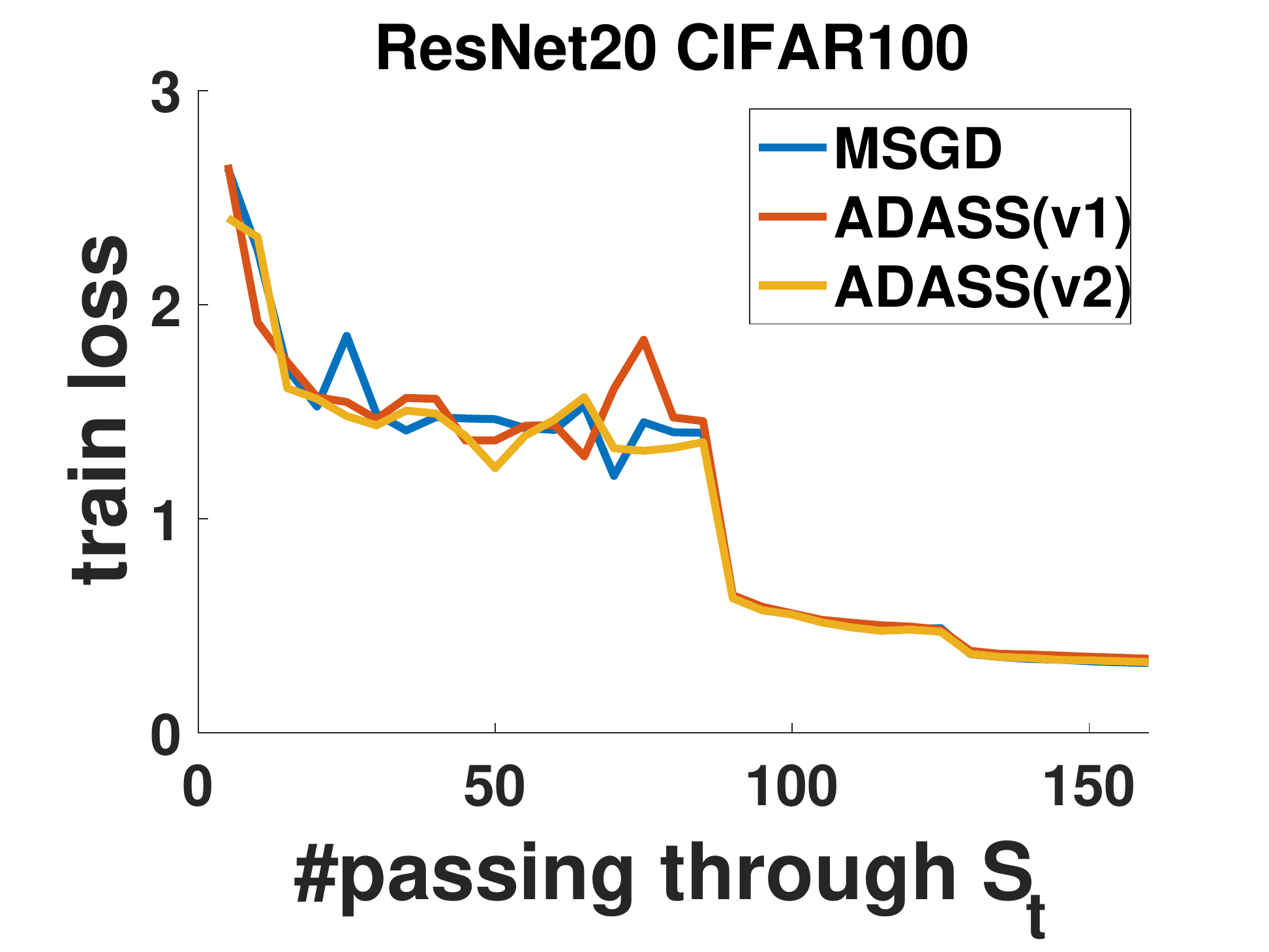}
  \includegraphics[width =3.5cm]{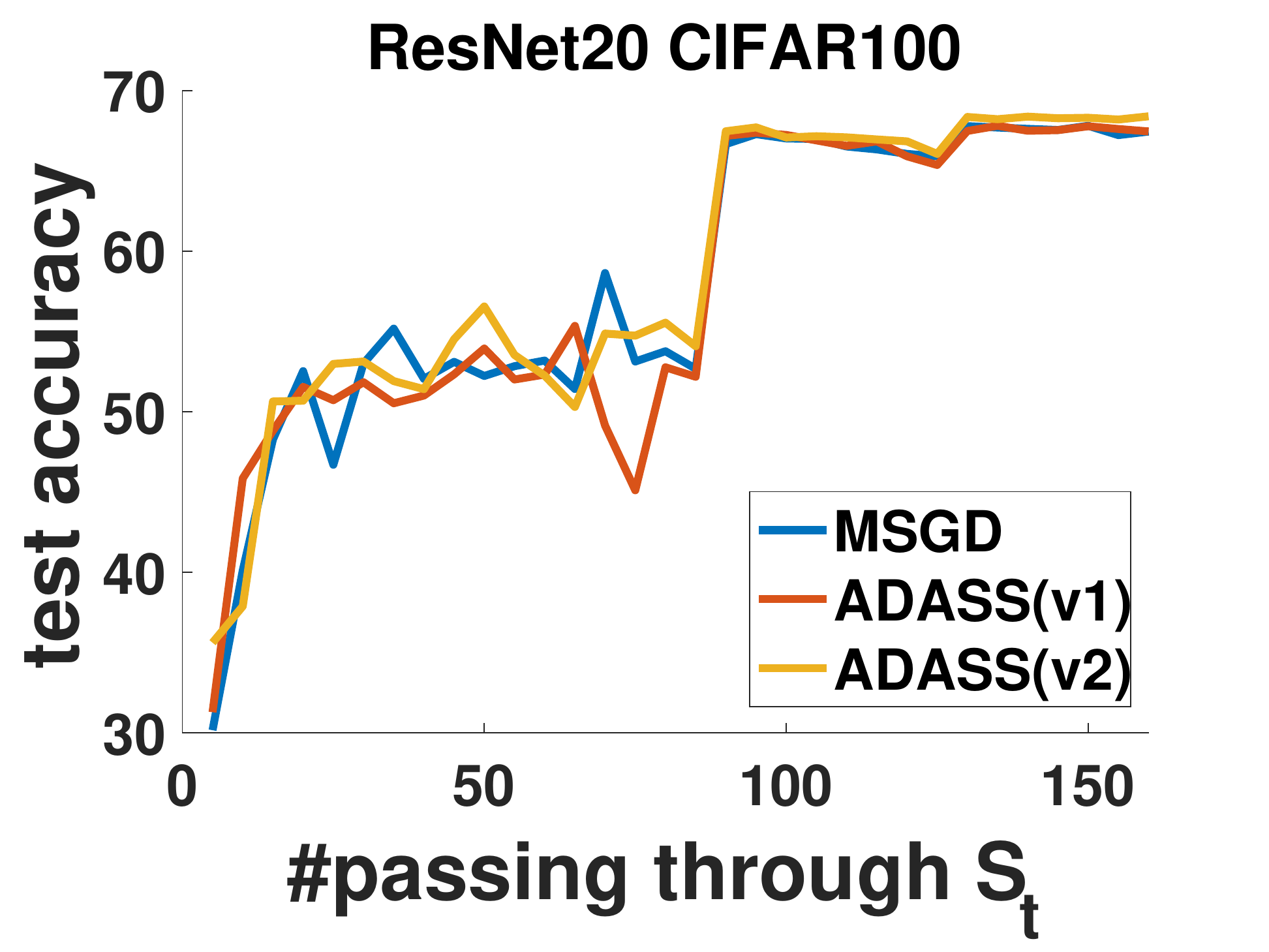}
  \includegraphics[width =3.5cm]{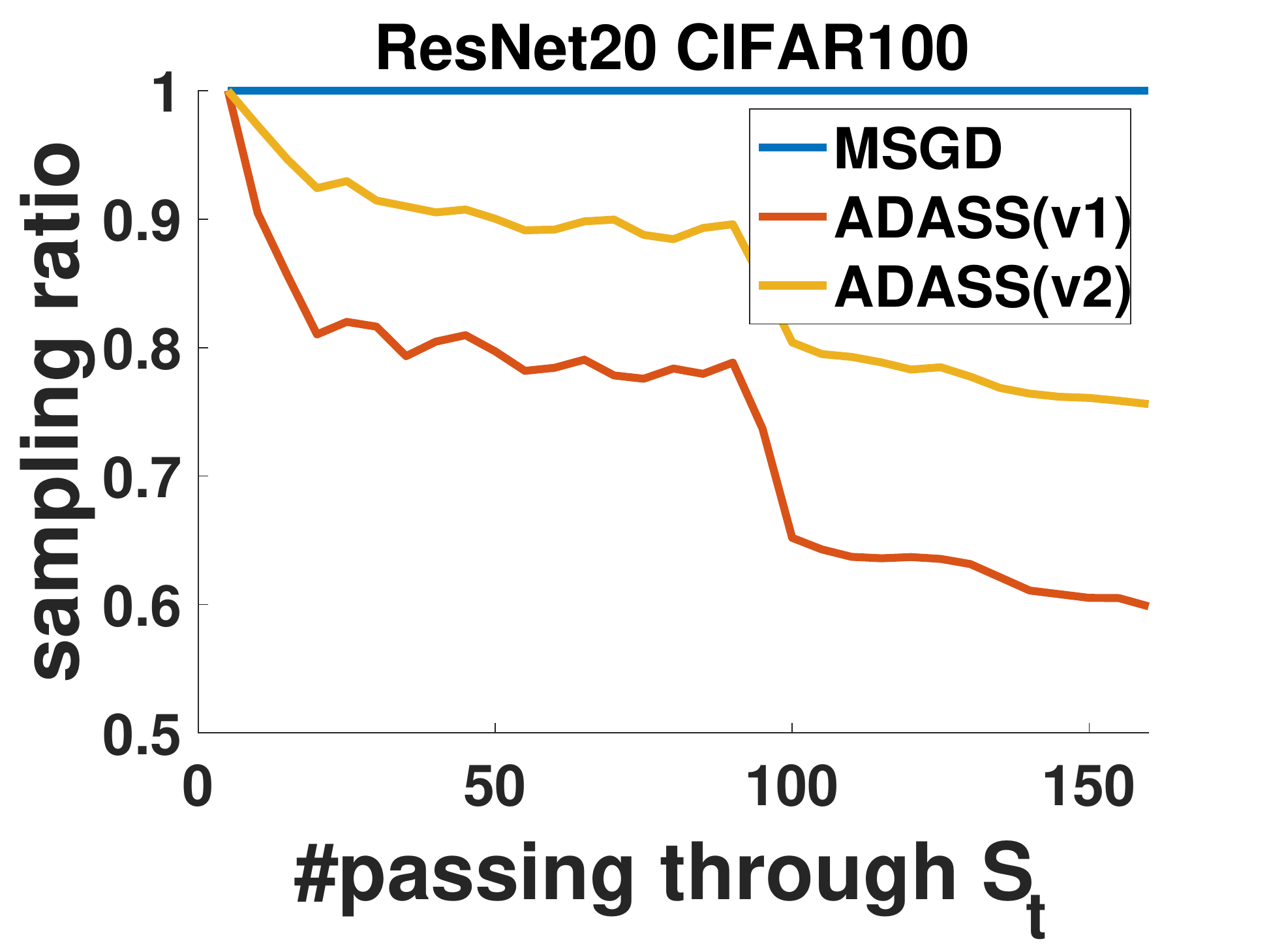}
  \includegraphics[width =3.5cm]{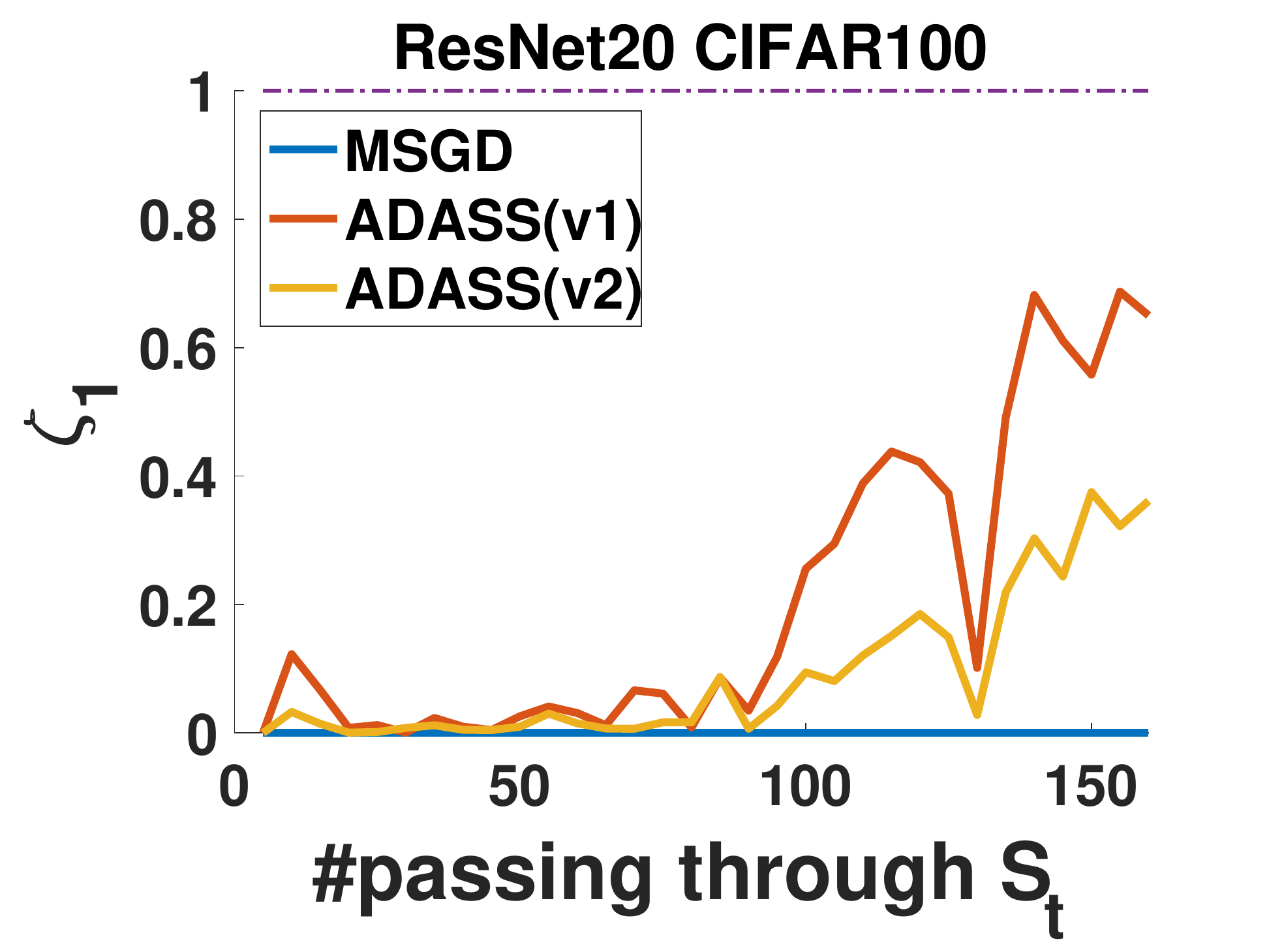}}
\subfigure{
  \includegraphics[width =3.5cm]{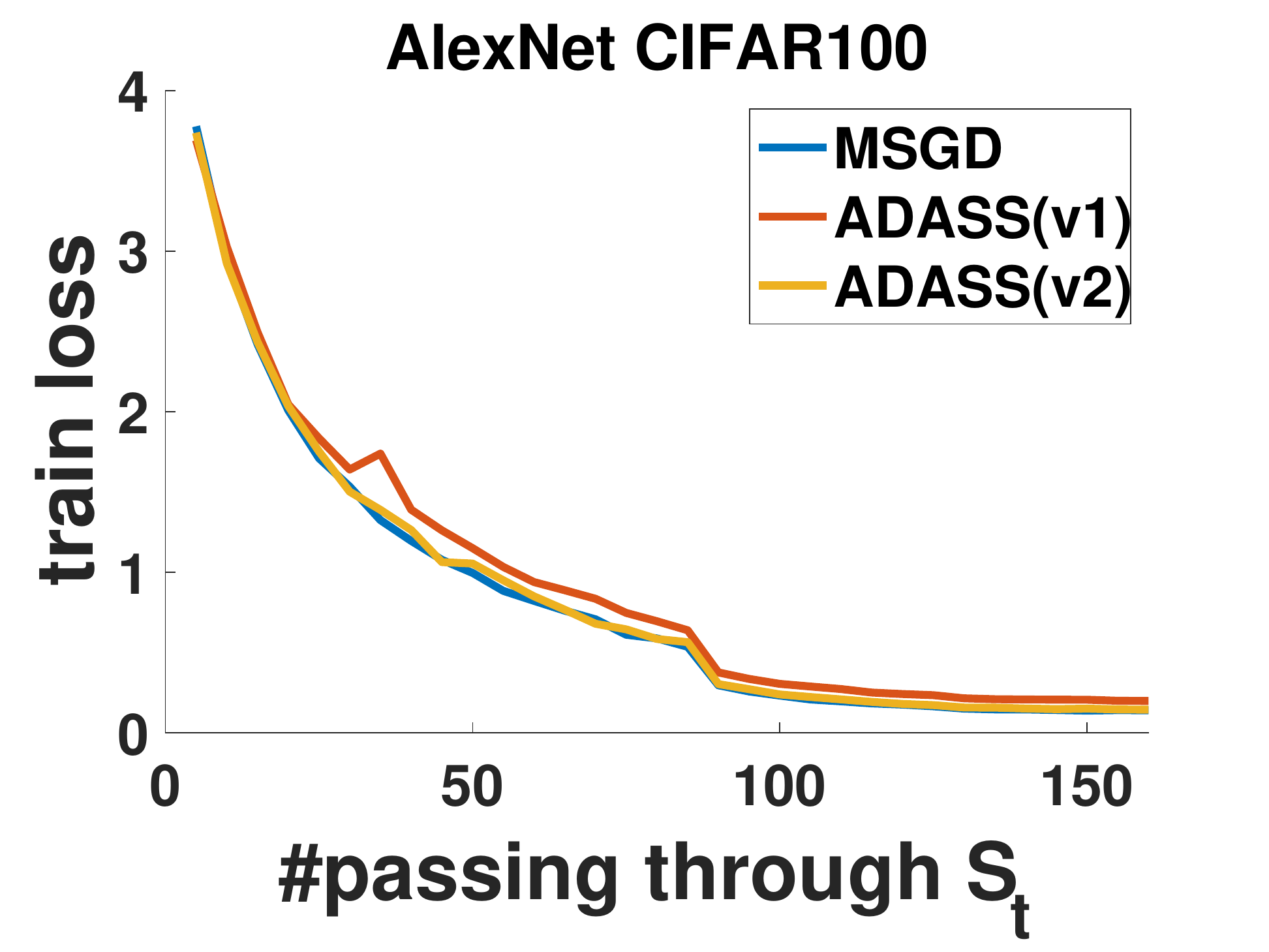}
  \includegraphics[width =3.5cm]{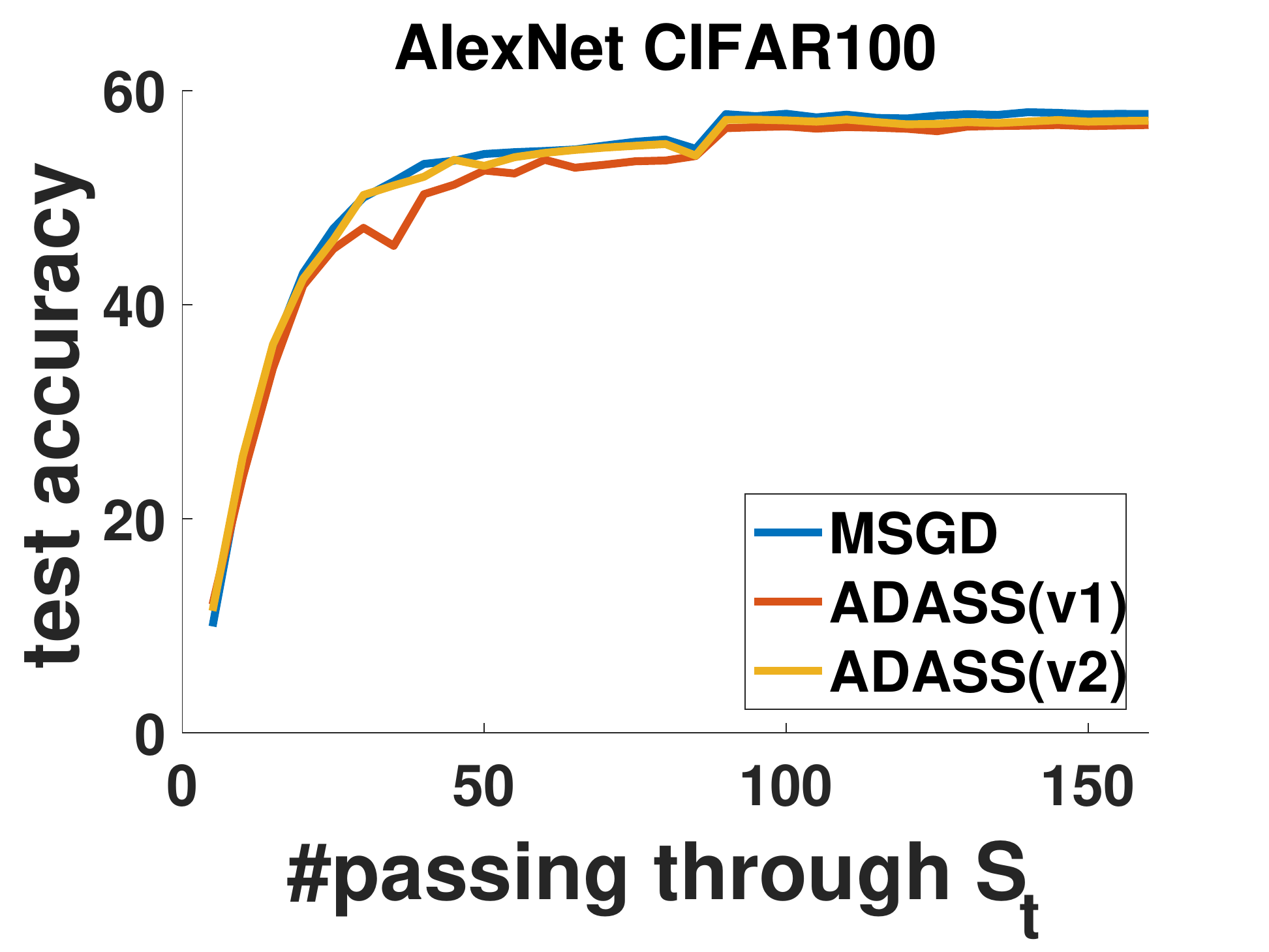}
  \includegraphics[width =3.5cm]{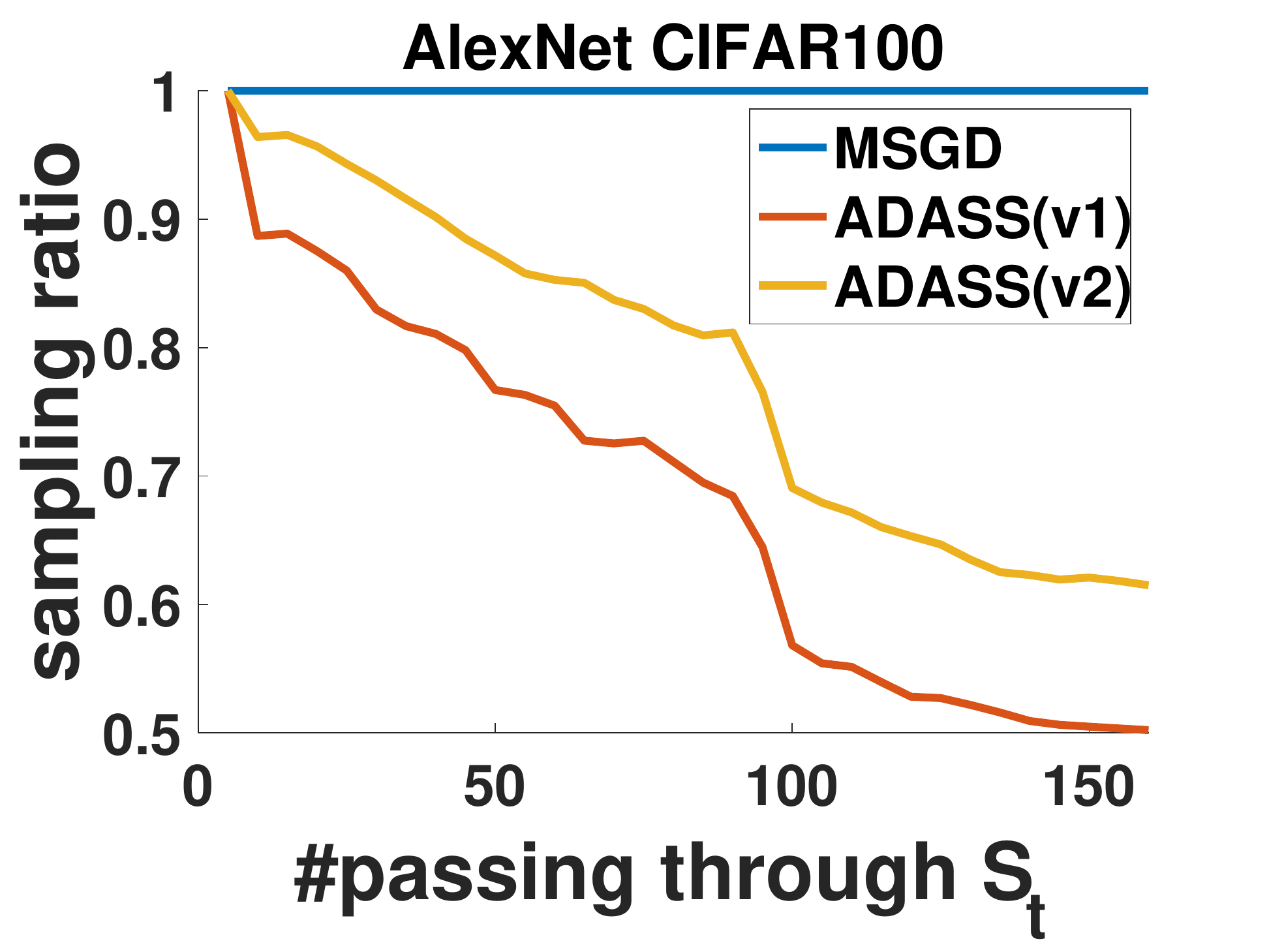}
  \includegraphics[width =3.5cm]{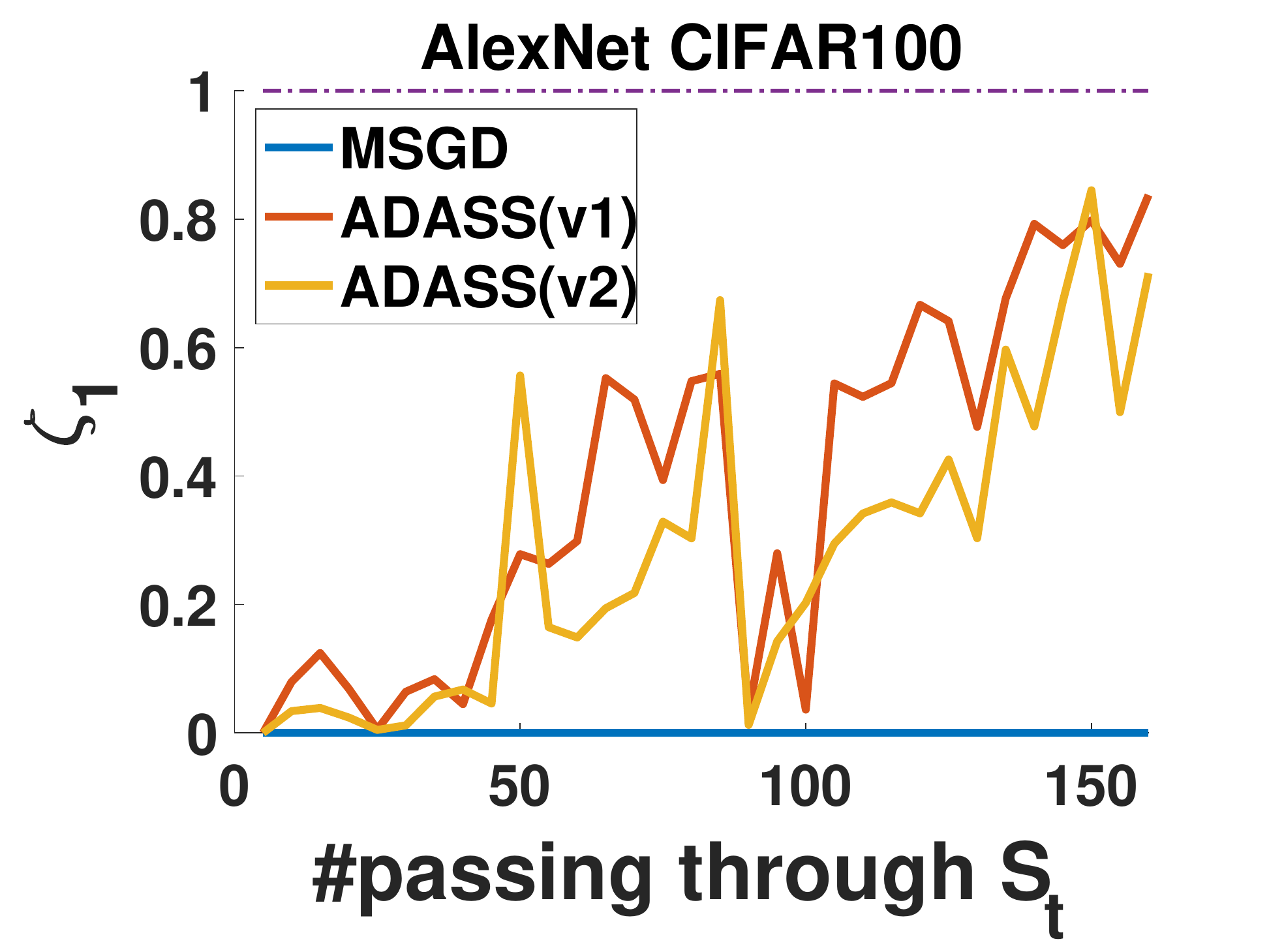}}
  \caption{Train ResNet20/AlexNet on CIFAR10/CIFAR100.}\label{exp:gamma_4}
\end{figure*}

First, we set $\gamma = \infty$ and $p=5$. The setting for $\gamma$ is common in recent deep learning training procedures. The results are presented in Figure~\ref{exp:gamma_infty}. The figures in the first column show the convergence result about training loss. We can see that ADASS get the same convergence result as MSGD in terms of passing through selected samples. The figures in the second column show the test accuracy. We can see that ADASS almost gets the same accuracy on test datasets. The figures in the third column show the sampling ratio, which is defined as $|S_t|/n$. $|S_t|/n \equiv 1$ in MSGD. When $\alpha<1$, the sampling ratio decreases growing with iterations. It implies that many training samples are useless and ADASS gets faster convergence rate. We also conduct experiments to verify the insignificance assumption that $\zeta_{1,t}<1$ in Theorem \ref{theorem:adass_converge1}. In each iteration, we pass 50 times through $S_t$ for searching the local minimum of $\hat{F}_{t}^\gamma$. The results are showing in the forth column. We can see that it is always less than 1. So the assumption is true in practice. At the same time, we calculate the $\rho_T$ defined in (\ref{eq:rho}), we can find that $\rho_T<1$ in the four experiments. This is consistent with previous result that ADASS gets faster convergence rate. We also set $\gamma = 10^4$ and the results are showing in Figure~\ref{exp:gamma_4}. We can find the similar phenomenons.

Next, we set different values for $p$ to compare the training wall clock time and test accuracy. The results are presented in Table~\ref{Tab}. We can see that $p$ does not have significant effect on training time and test accuracy. We can also get another interesting result from Figure~\ref{exp:gamma_infty}, Figure~\ref{exp:gamma_4} and Table~\ref{Tab}. In fact, the images in the two datasets are the same. The only difference is that each classification of CIFAR100 has less data than that of CIFAR10. In other word, data in CIFAR100 is more effective for classification task. So the sampling ratio of ADASS is higher on training CIFAR100.

\begin{table}
\centering
\caption{Empirical results with different hype parameters.}\label{Tab}
\begin{tabular}{cl|cc|cc}
    \hline
    ~                       & ~              & \multicolumn{2}{|c|}{CIFAR10} & \multicolumn{2}{|c}{CIFAR100}   \\ \hline
	Model                   & $(\alpha,p)$        & time      & test accuracy   & time      & test accuracy         \\ \hline
	\multirow{7}*{AlexNet}  & $(1,-)$        & 769       & 85.56\%         & 776       & 57.93\%               \\
	~                       & $(0.99,5)$     & 467       & 84.84\%         & 640       & 57.09\%               \\
	~                       & $(0.99,10)$    & 434       & 84.33\%         & 595       & 56.47\%               \\
	~                       & $(0.99,20)$    & 459       & 84.50\%         & 608       & 56.40\%               \\
	~                       & $(0.999,5)$    & 563       & 85.19\%         & 725       & 57.51\%               \\
	~                       & $(0.999,10)$   & 527       & 85.27\%         & 685       & 58.08\%               \\
	~                       & $(0.999,20)$   & 555       & 85.03\%         & 677       & 57.55\%               \\ \hline
    \multirow{7}*{ResNet20} & $(1,-)$        & 1126      & 91.61\%         & 1124      & 68.01\%               \\
	~                       & $(0.99,5)$     & 583       & 90.71\%         & 892       & 66.45\%               \\
	~                       & $(0.99,10)$    & 577       & 91.28\%         & 898       & 66.57\%               \\
	~                       & $(0.99,20)$    & 660       & 91.73\%         & 886       & 67.07\%               \\
	~                       & $(0.999,5)$    & 751       & 91.57\%         & 1093      & 67.76\%               \\
	~                       & $(0.999,10)$   & 744       & 91.65\%         & 1026      & 68.15\%               \\
	~                       & $(0.999,20)$   & 807       & 91.30\%         & 1025      & 67.65\%               \\
	\hline
\end{tabular}
\end{table}



\section{Conclusion}
In this paper, we propose a new method, called ADASS, for training acceleration. In ADASS, the sample size in each epoch of training can be smaller than the size of the full training set, by adaptively discarding some samples. ADASS can be seamlessly integrated with existing optimization methods, such as SGD and momentum SGD, for training acceleration. Empirical results show that ADASS can accelerate the training process of existing methods without sacrificing accuracy.

\bibliography{ref}

\begin{thebibliography}{28}
\providecommand{\natexlab}[1]{#1}
\providecommand{\url}[1]{\texttt{#1}}
\expandafter\ifx\csname urlstyle\endcsname\relax
  \providecommand{\doi}[1]{doi: #1}\else
  \providecommand{\doi}{doi: \begingroup \urlstyle{rm}\Url}\fi

\bibitem[Allen{-}Zhu(2018)]{DBLP:conf/icml/Allen-Zhu18}
Zeyuan Allen{-}Zhu.
\newblock Katyusha {X:} practical momentum method for stochastic
  sum-of-nonconvex optimization.
\newblock In \emph{Proceedings of the 35th International Conference on Machine
  Learning}, pages 179--185, 2018.

\bibitem[Borsos et~al.(2018)Borsos, Krause, and
  Levy]{DBLP:conf/colt/Borsos0L18}
Zalan Borsos, Andreas Krause, and Kfir~Y. Levy.
\newblock Online variance reduction for stochastic optimization.
\newblock In \emph{Conference On Learning Theory}, pages 324--357, 2018.

\bibitem[Bottou(2010)]{bottou-2010}
L\'{e}on Bottou.
\newblock Large-scale machine learning with stochastic gradient descent.
\newblock In \emph{Proceedings of the 19th International Conference on
  Computational Statistics}, 2010.

\bibitem[Chen et~al.(2019)Chen, Yuan, Yi, Zhou, Chen, and
  Yang]{chen2018universal}
Zaiyi Chen, Zhuoning Yuan, Jinfeng Yi, Bowen Zhou, Enhong Chen, and Tianbao
  Yang.
\newblock Universal stagewise learning for non-convex problems with convergence
  on averaged solutions.
\newblock In \emph{International Conference on Learning Representations}, 2019.

\bibitem[Csiba et~al.(2015)Csiba, Qu, and
  Richt{\'{a}}rik]{DBLP:conf/icml/CsibaQR15}
Dominik Csiba, Zheng Qu, and Peter Richt{\'{a}}rik.
\newblock Stochastic dual coordinate ascent with adaptive probabilities.
\newblock In \emph{Proceedings of the 32nd International Conference on Machine
  Learning}, pages 674--683, 2015.

\bibitem[Davis and Drusvyatskiy(2019)]{DBLP:journals/siamjo/DavisD19}
Damek Davis and Dmitriy Drusvyatskiy.
\newblock Stochastic model-based minimization of weakly convex functions.
\newblock \emph{{SIAM} Journal on Optimization}, 29\penalty0 (1):\penalty0
  207--239, 2019.

\bibitem[Defazio et~al.(2014)Defazio, Bach, and
  Lacoste{-}Julien]{DBLP:conf/nips/DefazioBL14}
Aaron Defazio, Francis~R. Bach, and Simon Lacoste{-}Julien.
\newblock {SAGA:} a fast incremental gradient method with support for
  non-strongly convex composite objectives.
\newblock In \emph{Advances in Neural Information Processing Systems}, pages
  1646--1654, 2014.

\bibitem[Duchi et~al.(2010)Duchi, Hazan, and Singer]{DBLP:conf/colt/DuchiHS10}
John~C. Duchi, Elad Hazan, and Yoram Singer.
\newblock Adaptive subgradient methods for online learning and stochastic
  optimization.
\newblock In \emph{{COLT} 2010 - The 23rd Conference on Learning Theory, Haifa,
  Israel, June 27-29, 2010}, pages 257--269, 2010.

\bibitem[Ghadimi and Lan(2016)]{DBLP:journals/mp/GhadimiL16}
Saeed Ghadimi and Guanghui Lan.
\newblock Accelerated gradient methods for nonconvex nonlinear and stochastic
  programming.
\newblock \emph{Math. Program.}, 156\penalty0 (1-2):\penalty0 59--99, 2016.

\bibitem[Hazan and Kale(2014)]{DBLP:journals/jmlr/HazanK14}
Elad Hazan and Satyen Kale.
\newblock Beyond the regret minimization barrier: optimal algorithms for
  stochastic strongly-convex optimization.
\newblock \emph{Journal of Machine Learning Research}, 15\penalty0
  (1):\penalty0 2489--2512, 2014.

\bibitem[Johnson and Zhang(2013)]{DBLP:conf/nips/Johnson013}
Rie Johnson and Tong Zhang.
\newblock Accelerating stochastic gradient descent using predictive variance
  reduction.
\newblock In \emph{Advances in Neural Information Processing Systems}, pages
  315--323, 2013.

\bibitem[Katharopoulos and Fleuret(2018)]{DBLP:conf/icml/KatharopoulosF18}
Angelos Katharopoulos and Fran{\c{c}}ois Fleuret.
\newblock Not all samples are created equal: Deep learning with importance
  sampling.
\newblock In \emph{Proceedings of the 35th International Conference on Machine
  Learning}, pages 2530--2539, 2018.

\bibitem[Kingma and Ba(2014)]{DBLP:journals/corr/KingmaB14}
Diederik~P. Kingma and Jimmy Ba.
\newblock Adam: A method for stochastic optimization.
\newblock \emph{CoRR}, abs/1412.6980, 2014.

\bibitem[Lan(2012)]{DBLP:journals/mp/Lan12}
Guanghui Lan.
\newblock An optimal method for stochastic composite optimization.
\newblock \emph{Math. Program.}, 133\penalty0 (1-2):\penalty0 365--397, 2012.

\bibitem[Leen and Orr(1993)]{DBLP:conf/nips/LeenO93}
Todd~K. Leen and Genevieve~B. Orr.
\newblock Optimal stochastic search and adaptive momentum.
\newblock In \emph{Advances in Neural Information Processing Systems}, pages
  477--484, 1993.

\bibitem[Lin et~al.(2017)Lin, Goyal, Girshick, He, and
  Doll{\'{a}}r]{DBLP:conf/iccv/LinGGHD17}
Tsung{-}Yi Lin, Priya Goyal, Ross~B. Girshick, Kaiming He, and Piotr
  Doll{\'{a}}r.
\newblock Focal loss for dense object detection.
\newblock In \emph{International Conference on Computer Vision}, pages
  2999--3007, 2017.

\bibitem[Namkoong et~al.(2017)Namkoong, Sinha, Yadlowsky, and
  Duchi]{DBLP:conf/icml/NamkoongSYD17}
Hongseok Namkoong, Aman Sinha, Steve Yadlowsky, and John~C. Duchi.
\newblock Adaptive sampling probabilities for non-smooth optimization.
\newblock In \emph{Proceedings of the 34th International Conference on Machine
  Learning}, pages 2574--2583, 2017.

\bibitem[Nesterov(2007)]{Nesterov07gradientmethods}
Yu. Nesterov.
\newblock Gradient methods for minimizing composite objective function, 2007.

\bibitem[Nitanda(2014)]{DBLP:conf/nips/Nitanda14}
Atsushi Nitanda.
\newblock Stochastic proximal gradient descent with acceleration techniques.
\newblock In \emph{Advances in Neural Information Processing Systems}, pages
  1574--1582, 2014.

\bibitem[Schmidt et~al.(2017)Schmidt, Roux, and
  Bach]{DBLP:journals/mp/SchmidtRB17}
Mark~W. Schmidt, Nicolas~Le Roux, and Francis~R. Bach.
\newblock Minimizing finite sums with the stochastic average gradient.
\newblock \emph{Math. Program.}, 162\penalty0 (1-2):\penalty0 83--112, 2017.

\bibitem[Shalev{-}Shwartz and
  Zhang(2013)]{DBLP:journals/jmlr/Shalev-Shwartz013}
Shai Shalev{-}Shwartz and Tong Zhang.
\newblock Stochastic dual coordinate ascent methods for regularized loss.
\newblock \emph{Journal of Machine Learning Research}, 14\penalty0
  (1):\penalty0 567--599, 2013.

\bibitem[Shalev{-}Shwartz and Zhang(2014)]{DBLP:conf/icml/Shalev-Shwartz014}
Shai Shalev{-}Shwartz and Tong Zhang.
\newblock Accelerated proximal stochastic dual coordinate ascent for
  regularized loss minimization.
\newblock In \emph{Proceedings of the 31th International Conference on Machine
  Learning}, pages 64--72, 2014.

\bibitem[Shrivastava et~al.(2016)Shrivastava, Gupta, and
  Girshick]{DBLP:conf/cvpr/ShrivastavaGG16}
Abhinav Shrivastava, Abhinav Gupta, and Ross~B. Girshick.
\newblock Training region-based object detectors with online hard example
  mining.
\newblock In \emph{Conference on Computer Vision and Pattern Recognition},
  pages 761--769, 2016.

\bibitem[Tseng(1998)]{DBLP:journals/siamjo/Tseng98}
Paul Tseng.
\newblock An incremental gradient(-projection) method with momentum term and
  adaptive stepsize rule.
\newblock \emph{{SIAM} Journal on Optimization}, 8\penalty0 (2):\penalty0
  506--531, 1998.

\bibitem[Xiao(2009)]{DBLP:conf/nips/Xiao09}
Lin Xiao.
\newblock Dual averaging method for regularized stochastic learning and online
  optimization.
\newblock In \emph{Advances in Neural Information Processing Systems}, pages
  2116--2124, 2009.

\bibitem[Zhang(2004)]{DBLP:conf/icml/Zhang04}
Tong Zhang.
\newblock Solving large scale linear prediction problems using stochastic
  gradient descent algorithms.
\newblock In \emph{Machine Learning, Proceedings of the Twenty-first
  International Conference}, 2004.

\bibitem[Zhao and Zhang(2015)]{DBLP:conf/icml/ZhaoZ15}
Peilin Zhao and Tong Zhang.
\newblock Stochastic optimization with importance sampling for regularized loss
  minimization.
\newblock In \emph{Proceedings of the 32nd International Conference on Machine
  Learning}, pages 1--9, 2015.

\bibitem[Zinkevich(2003)]{DBLP:conf/icml/Zinkevich03}
Martin Zinkevich.
\newblock Online convex programming and generalized infinitesimal gradient
  ascent.
\newblock In \emph{Machine Learning, Proceedings of the Twentieth International
  Conference}, pages 928--936, 2003.

\end{thebibliography}

\end{document}